\documentclass{article}

\PassOptionsToPackage{numbers,compress}{natbib}
\usepackage[preprint]{neurips_2026}

\usepackage[utf8]{inputenc} 
\usepackage[T1]{fontenc}    
\usepackage{hyperref}       
\usepackage{url}            
\usepackage{booktabs}       
\usepackage{amsfonts}       
\usepackage{nicefrac}       
\usepackage{microtype}      
\usepackage{xcolor}         
\usepackage{graphicx}
\usepackage{makecell}

\title{MonoVoc: Decoupling Geometry and Semantics for Lightweight Monocular Open-Vocabulary 3D Gaussians}

\author{Pouya Ardehkhani, Zahra Dehghanian, Morteza Abolghasemi, Hamid R. Rabiee\\
Sharif University of Technology, Tehran, Iran\\
{\tt\small \{pouya.ardehkhani02,zahra.dehghanian97, m.abolghasemi77, rabiee\}@sharif.edu}
}

\begin{document}

\maketitle

\begin{abstract}
Open vocabulary 3D scene understanding is essential for next-generation interactive systems, empowering users to intuitively query and navigate reconstructed environments using natural language. However, current 3D Gaussian frameworks are often bottlenecked by restrictive multiview capture requirements, costly scene-specific optimization, and the massive memory overhead of storing dense language features. We present a novel, training-free pipeline that fundamentally reimagines this paradigm by explicitly decoupling 3D geometric reconstruction from semantic integration. Given a standard monocular video sequence as input, our method efficiently outputs a compact, highly interpretable, and fully searchable object-level semantic Gaussian map. Rather than entangling heavy language embeddings within the mapping loop, we extract geometry independently and ground semantics through a lightweight, modular post-processing framework. Extensive evaluations on the Replica dataset demonstrate that this decoupled architecture preserves strong rendering fidelity and competitive segmentation accuracy. Crucially, by replacing dense per-Gaussian storage with modular, object-level semantic embeddings, our approach delivers an order-of-magnitude reduction in memory usage compared to SOTA baselines. This provides a highly efficient, scalable, and practical solution for open-vocabulary 3D retrieval and question answering directly from everyday monocular video.
\end{abstract}

\section{Introduction}

Open-vocabulary 3D scene understanding is important for interactive systems that must interpret real environments through natural language. In augmented reality~\cite{kato2023arkitscenerefer}, robotics~\cite{werby2024hovsg}, and inspection tasks~\cite{tasneem2026humanrobot}, users should be able to refer to objects with everyday descriptions, such as ``the red toolbox'' or ``the chair behind the table,'' without relying on a fixed category set.



Recent 3D Gaussian Splatting (3DGS)~\cite{kerbl20233dgaussians} methods have made 3D scenes searchable with language by attaching semantic or language features to reconstructed scenes. For example, SceneSplat~\cite{li2025scenesplat} learns open-vocabulary 3D features, OpenGaussian~\cite{wu2024opengaussian} builds instance-level semantic links, ObjectGS~\cite{zhu2025objectgs} brings object awareness to monocular Gaussian reconstruction, and LangSplatV2~\cite{li2025langsplatv2highdimensional3dlanguage} and 4D LangSplat~\cite{li20254dlangsplat4dlanguage} improve query efficiency or support dynamic scenes.

This line of work shows the promise of language-aware Gaussian maps, but practical monocular deployment also requires efficient reconstruction. Gaussian SLAM methods have made moving-camera 3D reconstruction increasingly practical \cite{matsuki2024gaussiansplatting,keetha2024splatam}, while open-vocabulary grounding and language-driven segmentation~\cite{li2022lseg} has enabled visual systems to respond to free-form language queries. Yet, in the monocular setting, this integration remains limited: semantics are often coupled with reconstruction, feature learning, or per-scene optimization, and dense semantic features are stored across many Gaussians. This makes the resulting maps costly, less editable, and harder to interpret as object-level representations. We therefore address a missing link in this landscape: a lightweight monocular 3DGS framework that produces maps that are not only renderable, but also searchable, segmentable, and queryable through compact open-vocabulary object semantics.


To address this gap, we propose a training-free open-vocabulary scene understanding and question answering framework built on a SLAM-generated 3DGS map. Given monocular video frames and segmentation masks, we reconstruct a geometry-aware Gaussian map and then add semantics as a post-processing step. First, we reverse the effect of alpha blending through our Semantic Color Deblending (SCD) algorithm to recover semantic color evidence for individual Gaussians. Second, we match the recovered colors to a fixed semantic palette, producing discrete object identities. Third, we assign one embedding to each object, so all Gaussians belonging to the same object share a compact semantic representation. This avoids storing a separate language feature for every Gaussian and enables efficient retrieval and question answering.




\section{Related Works}

Our work bridges the gap between 3D Gaussian Splatting (3DGS) and open-vocabulary language grounding. In particular, we focus on 3D representations that can be searched, segmented, and queried using free-form language. 
Existing approaches can be broadly categorized into two paradigms: multiview pipelines and monocular single-video pipelines. While both aim to bridge geometry and semantics, they differ fundamentally in capture constraints, reconstruction complexity, runtime efficiency, and the manner in which semantic information is encoded within the resulting 3D representation.

\subsection{Multiview language Gaussians}

Early open-vocabulary 3D grounding was shaped by LeRF~\cite{kerr2023lerf}, which learned dense language features on a NeRF radiance field~\cite{mildenhall2020nerf} and showed that text queries can be localized in 3D through rendered relevancy maps. However, LeRF depends on well-covered, calibrated multiview captures, requires costly optimization, and stores semantics densely, making it less suitable for long monocular videos.

Recent Gaussian-based methods improve efficiency and scalability. LangSplatV2~\cite{li2025langsplatv2highdimensional3dlanguage} enables faster rendering of high-dimensional language features, while 4D LangSplat~\cite{li20254dlangsplat4dlanguage} extends language-aware Gaussians to dynamic scenes. Other methods improve semantic structure and boundary quality: OpenGaussian~\cite{wu2024opengaussian} uses 2D masks and 2D--3D links, LangSurf~\cite{li2024langsurf} aligns language features with object surfaces, and systems such as SLGaussian~\cite{chen2025slgaussian} and SLAG~\cite{szilagyi2025slag} attach language features to Gaussian maps.



Together, these methods achieve strong open-vocabulary grounding, faster querying, and sharper segmentation in multiview or well-controlled reconstruction settings. However, they are less suited to our target setting: a single monocular walkthrough. In this case, the input is less complete, camera coverage is more limited, and optimization-heavy semantic learning or dense feature storage becomes costly and harder to scale. As a result, the performance and practicality of these methods do not directly carry over to ordinary monocular videos. In contrast, our approach adds semantics after reconstruction and produces a lightweight, compact, and interpretable object-level map.

\subsection{Monocular open-vocabulary 3DGS understanding}

Monocular methods are closer to realistic capture scenarios, where a user records a scene with one moving camera. However, existing approaches often still couple semantic understanding with reconstruction and store semantic information densely inside the Gaussian representation.

ObjectGS~\cite{zhu2025objectgs} moves toward object-centric monocular Gaussian reconstruction. It models objects using local anchors that generate associated Gaussians while preserving object identity. This improves object separation and supports more object-level reasoning. However, reconstruction and semantic modeling remain tightly coupled in an optimization-heavy pipeline, and the semantic state is not necessarily lightweight or easy to inspect.

SceneSplat~\cite{li2025scenesplat} takes a different direction by using vision--language pretraining and large-scale training to learn generalizable 3D features on Gaussian representations. This is useful for open-vocabulary scene understanding, but it introduces high training cost and does not directly provide a simple end-to-end monocular reconstruction-and-query pipeline. Its output is also still mainly feature-based rather than an explicitly readable object-level semantic map.

Our work is closest in spirit to ObjectGS~\cite{zhu2025objectgs} and SceneSplat~\cite{li2025scenesplat}, since both address semantic understanding for 3DGS in settings closer to monocular capture. However, their efficiency does not directly match our target setting. ObjectGS couples object semantics with reconstruction and optimization, while SceneSplat predicts dense open-vocabulary features over 3DGS representations and does not include 3D map generation in its reported runtime. These design choices lead to larger maps, higher memory cost, or less direct end-to-end runtime comparison. To the best of our knowledge, we are the first to study a training-free, SLAM-based monocular 3DGS pipeline that attaches compact object-level open-vocabulary semantics after reconstruction, without scene-specific semantic optimization or dense per-Gaussian language storage. By assigning one embedding per object rather than one feature per Gaussian, MonoVoc produces a compact, object-readable semantic map for efficient retrieval and question answering.



\section{Proposed Method}

\begin{figure}[h]
  \centering
  \includegraphics[width=\linewidth]{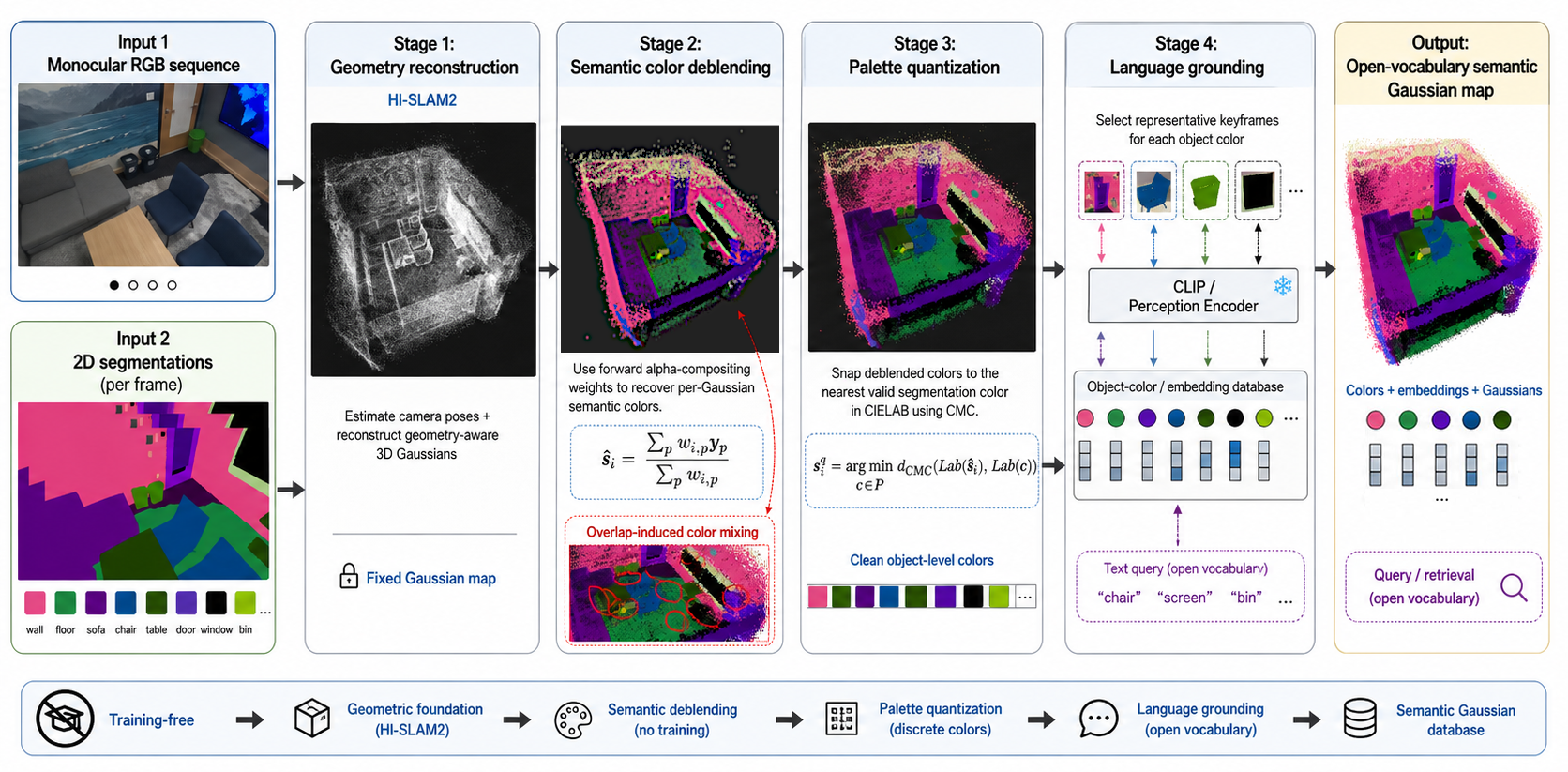}
  \caption{
  Overview of the proposed training-free open-vocabulary semantic Gaussian mapping pipeline. Given a monocular RGB video, we first reconstruct a geometry-aware 3D Gaussian map using HI-SLAM2~\cite{zhang2024hislam2}. We then keep the geometry fixed and add semantics as a post-processing step. 2D segmentation masks provide semantic color observations, which are assigned to Gaussians through Semantic Color Deblending (SCD). The recovered colors are then snapped to a discrete segmentation palette to obtain object-level identities. For each object, representative masked keyframes are used to extract language embeddings. The final database links Gaussians, semantic object colors, and embeddings, enabling open-vocabulary retrieval and question answering. Additional discussion of overlap in deblending is provided in Appendix~\ref{app:overlap}.
  }
  \label{fig:semantic-gaussian-mapping-overview}
\end{figure}

Figure~\ref{fig:semantic-gaussian-mapping-overview} summarizes our training-free pipeline for open-vocabulary semantic Gaussian mapping. Given a monocular RGB video, we first reconstruct a geometry-aware 3D Gaussian map with HI-SLAM2~\cite{zhang2024hislam2}, then keep the geometry fixed and add semantics as post-processing. 2D segmentation masks provide semantic color observations, which are assigned to Gaussians using Semantic Color Deblending (SCD); the recovered colors are then quantized to the segmentation palette to obtain object-level identities. Finally, representative masked keyframes are used to extract language embeddings for each object color. The resulting compact database links Gaussians, object identities, and embeddings for open-vocabulary retrieval and question answering. We next detail the pipeline components: alpha compositing, SCD, palette quantization, and language embedding assignment.

\subsection{Training-free open-vocabulary semantic mapping}
\label{sec:proposed_pipeline}



Our method is a SLAM-based, training-free pipeline for open-vocabulary 3D Gaussian understanding from monocular videos, as shown in Figure~\ref{fig:semantic-gaussian-mapping-overview}. Its key idea is to decouple geometry reconstruction from semantic grounding: we first build a stable Gaussian map from RGB input, then attach object colors and language embeddings. This avoids retraining the SLAM system or optimizing language features inside the mapping loop, while producing a compact semantic map queryable with natural language.

The pipeline has four stages. First, we reconstruct the scene with HI-SLAM2~\cite{zhang2024hislam2}, a geometry-aware Gaussian SLAM system for fast monocular RGB reconstruction. HI-SLAM2 combines monocular geometry priors with dense SLAM and represents the map using 3D Gaussian splatting, yielding a renderable, geometrically meaningful scene representation. We use this Gaussian map as fixed geometric support for semantic assignment. Section~\ref{sec:forward_rendering} reviews the forward alpha-compositing process used by the map.

Second, we recover an initial semantic RGB color for each Gaussian using our
Semantic Color Deblending (SCD) algorithm. Given 2D segmentation maps, which can
come from any reliable segmentation source such as SAM~\cite{kirillov2023segmentanything},
and the fixed Gaussian geometry, SCD uses the alpha-compositing contribution of
each Gaussian to collect semantic evidence from all pixels it helps render. The
resulting per-Gaussian estimate is
\begin{equation}
\hat{s}_i=
\frac{\sum_p w_{i,p}y_p}{\sum_p w_{i,p}},
\label{eq:pipeline_deblend}
\end{equation}
where $w_{i,p}$ is the contribution of Gaussian $i$ at pixel observation $p$,
and $y_p$ is the background-subtracted semantic color. Intuitively, this step
reverses the renderer: instead of asking what color a Gaussian produces at a
pixel, we ask which semantic color best explains all pixels influenced by that
Gaussian. The derivation, exact-recovery condition, and overlap-bias analysis of
SCD are provided in Section~\ref{sec:semantic_deblending}.

Third, we correct the deblended colors using palette quantization. Because
alpha blending can mix colors from overlapping Gaussians, the recovered color
$\hat{s}_i$ may lie between valid 2D segmentation colors. We therefore extract
the discrete object palette $\mathcal{P}$ from the segmented images and assign
each Gaussian to the closest valid semantic color. To make this comparison
perceptually meaningful, we compare colors in CIELAB
space~\cite{standard2007colorimetry} using CMC distance~\cite{clarke1984modification}:
\begin{equation}
s_i^{q}
=
{\rm arg\,min}_{c\in\mathcal{P}}
d_{\mathrm{CMC}}
\left(
\mathrm{Lab}(\hat{s}_i),
\mathrm{Lab}(c)
\right).
\label{eq:pipeline_quantization}
\end{equation}
This step snaps noisy or blended Gaussian colors back to the object-level labels
observed in the 2D segmentations. Additional explanation of why overlap appears
in Gaussian deblending is provided in Appendix~\ref{app:overlap}.


Finally, we attach language embeddings to object colors. For each semantic RGB color, we select representative keyframes based on its prevalence across segmented frames, providing clean object views for embedding extraction. The embedding backend is modular: masked regions can be captioned with a vision-language model such as Osprey~\cite{yuan2024osprey} and embedded with a language model, or directly encoded with alignment models such as CLIP~\cite{radford2021learning}, DINO~\cite{caron2021emerging}, or Perception Encoder~\cite{bolya2025PerceptionEncoder}. This masked-region design relates to open-vocabulary segmentation methods such as OVSeg~\cite{liang2023ovseg}, which classify mask proposals with vision-language models. In our implementation, we use CLIP and Perception Encoder with hard- and soft-mask variants, store each semantic-color embedding in a database, and link it to all Gaussians assigned that color.

Our contribution is therefore a modular monocular pipeline for open-vocabulary
3DGS understanding. Geometry comes from HI-SLAM2, object identity comes from
deblended and quantized segmentation colors, and language meaning comes from
masked keyframe embeddings. This produces a compact and readable semantic
Gaussian map without per-scene semantic training, while supporting efficient
retrieval, question answering, and command grounding in reconstructed 3D scenes.

\subsection{Forward Rendering Equation}
\label{sec:forward_rendering}

We first describe the forward rendering process that our method later inverts.
For each pixel $p$, the projected Gaussians are sorted from front to back and
alpha-composited into a final semantic color. Each Gaussian $i$ has a semantic
color $\mathbf{s}_i$ (not RGB value rather a learnable parameter), opacity $o_i$, projected mean $\mu_i$, and screen-space
covariance $\Sigma_i$. Its pixel-level opacity is computed as
\begin{equation}
\alpha_i(p)=\min\left(0.99,\;
o_i\exp\left[-\frac{1}{2}
(p-\mu_i)^{\top}
\Sigma_i^{-1}
(p-\mu_i)\right]\right).
\label{eq:alpha_computation}
\end{equation}
As shown in Equation~\ref{eq:alpha_computation}, the contribution is high near
the Gaussian center and decreases smoothly according to the projected covariance.
Occlusion from closer Gaussians is captured by the accumulated transmittance:
\begin{equation}
T_i(p)=\prod_{j<i}\left(1-\alpha_j(p)\right),
\qquad
w_i(p)=\alpha_i(p)T_i(p).
\label{eq:transmittance_weight}
\end{equation}
Equation~\ref{eq:transmittance_weight} defines how much visibility remains
before Gaussian $i$ and its effective weight $w_i$ at pixel $p$. The final
rendered semantic color is then
\begin{equation}
\mathbf{C}(p)=\sum_{i=1}^{N} w_i(p)\mathbf{s}_i
+T_N(p)\mathbf{b},
\label{eq:forward_rendering}
\end{equation}
where $\mathbf{b}$ is the background color and $T_N$ is the remaining
transmittance after all Gaussians. Equation~\ref{eq:forward_rendering} makes
the key structure explicit: each pixel is a weighted mixture of Gaussian semantic
colors. Our inverse formulation uses these same weights to recover the underlying
semantic colors from rendered observations.

\subsection{Semantic Color Deblending (SCD)}
\label{sec:semantic_deblending}

We now derive the Semantic Color Deblending (SCD) algorithm used to assign one semantic color to each Gaussian. Let $p$ index a pixel observation, or a pixel--frame observation in
the multi-view setting. For fixed Gaussian geometry, opacity, and depth ordering,
the compositing weights are known, and the only unknowns are the Gaussian
semantic colors $s_i$. The forward semantic rendering equation is
\begin{equation}
C_p=\sum_{i=1}^{N} w_{i,p}s_i+T_{N+1,p}b,
\qquad
w_{i,p}=\alpha_{i,p}T_{i,p},
\label{eq:deblend_forward_affine}
\end{equation}
where $b$ is the background semantic color. Since $w_{i,p}$ and $T_{N+1,p}$ do
not depend on $s_i$, the renderer is affine in the semantic colors. After
subtracting the known background term,
$y_p=C_p-T_{N+1,p}b$,
we obtain the linear inverse problem
\begin{equation}
y_p=\sum_{i=1}^{N} w_{i,p}s_i.
\label{eq:deblend_linear_system}
\end{equation}

\paragraph{Theorem 1 (Compositing weights are exact mixing coefficients).}
\emph{Statement.}
Let
$T_{i,p}=\prod_{j<i}\left(1-\alpha_{j,p}\right)$
and 
$w_{i,p}=\alpha_{i,p}T_{i,p}$.
Then
\begin{equation}
w_{i,p}\ge 0,
\qquad
\sum_{i=1}^{N}w_{i,p}+T_{N+1,p}=1.
\label{eq:deblend_convex_sum}
\end{equation}
Thus each pixel is an exact convex mixture of Gaussian semantic colors and the
background.

\emph{Proof.}
Since $\alpha_{i,p}\in[0,1)$ and $T_{i,p}$ is a product of nonnegative terms,
$w_{i,p}\ge 0$. Also,
$T_{i+1,p}=T_{i,p}\left(1-\alpha_{i,p}\right)$,
so
\begin{equation}
w_{i,p}
=
\alpha_{i,p}T_{i,p}
=
T_{i,p}-T_{i+1,p}.
\label{eq:deblend_telescoping_step}
\end{equation}
Summing Equation~(\ref{eq:deblend_telescoping_step}) over all Gaussians gives
\begin{equation}
\sum_{i=1}^{N}w_{i,p}
=
T_{1,p}-T_{N+1,p}
=
1-T_{N+1,p},
\end{equation}
which proves Equation~(\ref{eq:deblend_convex_sum}).

\paragraph{Theorem 2 (Weighted-average deblending).}
\emph{Statement.}
For each Gaussian $i$, consider the per-Gaussian weighted least-squares problem
$\hat{s}_i={\rm arg\,min}_{s\in\mathbb{R}^3}\sum_p w_{i,p}\left\|s-y_p\right\|_2^2$.
If
$W_i=\sum_p w_{i,p}>0$, 
then the unique minimizer is
\begin{equation}
\hat{s}_i=
\frac{\sum_p w_{i,p}y_p}{\sum_p w_{i,p}}.
\label{eq:deblend_weighted_average}
\end{equation}

\emph{Proof.}
Let
$J_i(s)=\sum_p w_{i,p}\left\|s-y_p\right\|_2^2$.
Differentiating with respect to $s$ gives
\begin{equation}
\frac{\partial J_i}{\partial s}
=
2\left(\sum_p w_{i,p}\right)s
-
2\sum_p w_{i,p}y_p.
\label{eq:deblend_gradient}
\end{equation}
Setting Equation~(\ref{eq:deblend_gradient}) to zero gives
\begin{equation}
\left(\sum_p w_{i,p}\right)\hat{s}_i
=
\sum_p w_{i,p}y_p.
\end{equation}
Because $W_i>0$, division by $W_i$ gives
Equation~(\ref{eq:deblend_weighted_average}). The objective is a strictly convex
quadratic when $W_i>0$, so the minimizer is unique.

\paragraph{Theorem 3 (Exact recovery in the non-overlapping case).}
\emph{Statement.}
Assume each pixel is explained by exactly one Gaussian. That is, for each $p$
there exists an index $\iota(p)$ such that
\begin{equation}
w_{\iota(p),p}=1,
\qquad
w_{j,p}=0
\quad
\mathrm{for}
\quad
j\ne \iota(p).
\label{eq:deblend_one_hot}
\end{equation}
Then the estimator in Equation~(\ref{eq:deblend_weighted_average}) recovers the
true semantic color exactly: $\hat{s}_i=s_i$ for every Gaussian with $W_i>0$.

\emph{Proof.}
Using Equation~(\ref{eq:deblend_linear_system}) and the one-hot condition in
Equation~(\ref{eq:deblend_one_hot}), each observation satisfies
$y_p=s_{\iota(p)}$.
For Gaussian $i$, only pixels with $\iota(p)=i$ have nonzero weight. Therefore,
\begin{equation}
\hat{s}_i
=
\frac{\sum_{p:\iota(p)=i} y_p}{\sum_{p:\iota(p)=i}1}
=
\frac{\sum_{p:\iota(p)=i} s_i}{\sum_{p:\iota(p)=i}1}
=
s_i.
\end{equation}
Thus the estimator is exact when Gaussian contributions do not overlap.

\paragraph{Theorem 4 (Overlap bias of weighted-average deblending).}
\emph{Statement.}
Assume the linear model in Equation~(\ref{eq:deblend_linear_system}). Define
$G_{ij}=\sum_p w_{i,p}w_{j,p}$,
and
$W_i=\sum_p w_{i,p}$.
If $W_i>0$, then the estimator in
Equation~(\ref{eq:deblend_weighted_average}) satisfies
\begin{equation}
\hat{s}_i
=
\frac{G_{ii}}{W_i}s_i
+
\sum_{j\ne i}
\frac{G_{ij}}{W_i}s_j.
\label{eq:deblend_bias}
\end{equation}

\emph{Proof.}
Substituting Equation~(\ref{eq:deblend_linear_system}) into
Equation~(\ref{eq:deblend_weighted_average}) gives
\begin{equation}
\hat{s}_i
=
\frac{1}{W_i}
\sum_p w_{i,p}
\left(\sum_{j=1}^{N}w_{j,p}s_j\right).
\end{equation}
Rearranging the sums gives
\begin{equation}
\hat{s}_i
=
\frac{1}{W_i}
\sum_{j=1}^{N}
\left(\sum_p w_{i,p}w_{j,p}\right)s_j
=
\frac{1}{W_i}
\sum_{j=1}^{N}G_{ij}s_j.
\end{equation}
Separating the term $j=i$ gives Equation~(\ref{eq:deblend_bias}). Therefore the
bias comes only from overlap with other Gaussians, measured by the off-diagonal
terms $G_{ij}$.

Equations~(\ref{eq:deblend_weighted_average}) and
(\ref{eq:deblend_bias}) summarize the proposed deblending step: it is the exact
minimizer of a per-Gaussian weighted least-squares problem, it is exact for
one-hot visibility, and under overlap it has an explicit mixing bias. We include
only the main derivations here for clarity; additional details and full proofs
are provided in Appendix~\ref{app:deblending_proofs}.

\section{Experiment}
We evaluate the proposed pipeline in terms of reconstruction quality, semantic color quantization, language-based object retrieval, and map efficiency: including memory, runtime, and number of Gaussians. Further analyses and additional experiments are provided in Appendix~\ref{app:FEX}.

\subsection{Experiment Setup}
\label{sec:experimental_setup}


We evaluate on Replica~\cite{straub2019replica} using monocular RGB sequences as input and the corresponding semantic segmentations as 2D supervision for our training-free semantic mapping pipeline. For each scene, we reconstruct a 3D Gaussian map with HI-SLAM2~\cite{zhang2024hislam2} using its fixed Replica configuration; our method uses no additional training, optimizer, or train/test split. We then apply semantic deblending and palette quantization with identical settings across scenes: batch size 32 for extracting unique colors from ground-truth segmentations, Stage 2 deblending with 9000 Gaussians per batch at $616\times344$ segmentation resolution, and Stage 3 palette quantization using CMC distance in Lab space with $l=2.0$, $c=1.0$, and GPU batch size 100,000. For end-to-end semantic rendering error, we render the colored Gaussian point cloud with point size 0.05 for measuring error in stage 3. All experiments are single-run evaluations on one NVIDIA A100 GPU.

\subsection{Reconstruction and Rendered Segmentation Evaluation}
\label{sec:stage1_evaluation}


In the first stage, we assess whether the reconstructed Gaussian map preserves visual quality and semantic consistency after rendering. We report PSNR~\cite{hore2010universal}, SSIM~\cite{wang2004ssim}, and LPIPS~\cite{zhang2018lpips} for rendering quality, and mIoU~\cite{everingham2010pascal} for rendered mask accuracy. As shown in Table~\ref{tab:stage1_results}, our method achieves the best PSNR and LPIPS, indicating sharper and more perceptually faithful renderings. While SceneSplat attains the highest mIoU, our method is close without scene-specific semantic training or dense per-Gaussian language storage. These results show that HI-SLAM2 provides a strong geometric backbone, and that our post-processing semantic assignment maintains competitive segmentation quality in a lighter, more modular pipeline.


\begin{table}[ht]
  \caption{MonoVoc's Stage 1 reconstruction and rendered segmentation results on Replica. We compare with ObjectGS~\cite{zhu2025objectgs} and SceneSplat~\cite{li2025scenesplat}, the closest monocular 3D Gaussian-based methods to our setting. Reported variation is the standard deviation over multiple scenes.}
  \label{tab:stage1_results}
  \centering
  \begin{tabular}{lcccc}
    \toprule
    Model & PSNR $\uparrow$ & SSIM $\uparrow$ & LPIPS $\downarrow$ & mIoU $\uparrow$ \\
    \midrule
    ObjectGS~\cite{zhu2025objectgs} 
    & $40.26 \pm 0.32$ 
    & $\mathbf{0.984} \pm 0.0015$ 
    & $0.028 \pm 0.0021$ 
    & $88.39 \pm 0.78\%$ \\

    SceneSplat~\cite{li2025scenesplat} 
    & $41.25 \pm 0.35$ 
    & $0.980 \pm 0.0018$ 
    & $0.040 \pm 0.0025$ 
    & $\mathbf{91.89} \pm 0.84\%$ \\

    MonoVoc (ours) 
    & $\mathbf{41.91} \pm 0.34$ 
    & $0.981 \pm 0.0012$ 
    & $\mathbf{0.025} \pm 0.0016$ 
    & $91.22 \pm 0.72\%$ \\
    \bottomrule
  \end{tabular}
\end{table}

\subsection{Quantization Metric Selection}
\label{sec:stage3_quantization_results}

In Stage 3, the deblended Gaussian colors are mapped back to the nearest valid color in the semantic palette. Because alpha blending and overlapping Gaussians can slightly shift these colors, the distance metric used for palette snapping has a direct impact on the final segmentation quality. We compare six quantization metrics: CAM16~\cite{li2017cam16}, CIE94~\cite{mcdonald1995cie94}, CIEDE2000~\cite{luo2001ciede2000}, CMC~\cite{clarke1984modification}, DIN99o~\cite{cui2002din99}, and Euclidean distance in RGB space. Each metric is evaluated using a pixel-level semantic mismatch score over 2000 rendered semantic images.

\begin{figure}[h]
  \centering
  \includegraphics[width=0.55\linewidth, height=0.45\linewidth]{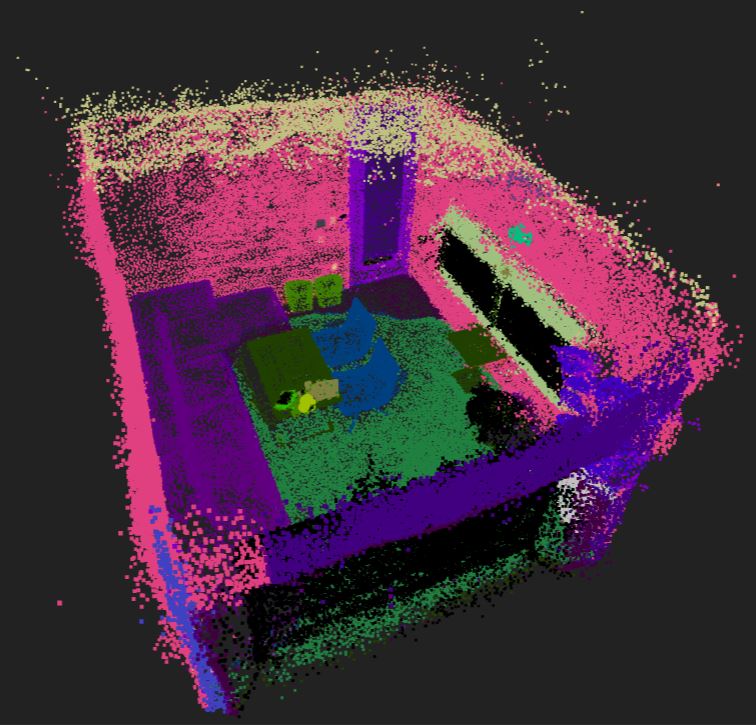}
  \caption{CMC~\cite{clarke1984modification}-based palette quantization snaps deblended colors to valid semantic colors.}
  \label{fig:cmc_quantization}
\end{figure}

To generate each rendered image, we project the colored 3D point cloud into the camera view using the estimated pose and intrinsics, remove points behind the camera, depth-sort the visible points, and draw them on a 2D canvas as colored circles or squares with a point size of 0.05. The rendered segmentation is then compared with the reference segmentation after nearest-neighbor alignment to avoid interpolation artifacts and preserve semantic colors. Pure-white target pixels are ignored since they represent unlabeled or empty regions, and we report errors both over all pixels and over non-white pixels only. As shown in Table~\ref{tab:quantization_metric_results}, CMC achieves the fewest mismatched pixels and the lowest error rates, reducing the non-white error to 7.48\%. This suggests that CMC is the most reliable choice for our palette snapping step, as it handles small perceptual color shifts from deblending while still assigning Gaussians to stable object-level semantic colors, as illustrated in Figure~\ref{fig:cmc_quantization}. Therefore, we use CMC as the default quantization distance.

\begin{table}[ht]
  \caption{Comparison of color-distance metrics for Stage 3 palette quantization over 2000 rendered semantic images and 10,566,418,000 total pixels. Reported variation is the std over multiple scenes.}
  \label{tab:quantization_metric_results}
  \centering
  \begin{tabular}{lcccc}
    \toprule
    Method & \makecell{Mismatched\\Pixels $\downarrow$} & \makecell{Mismatch\\Ratio $\downarrow$} & \makecell{Overall Err.\\(all) $\downarrow$} & \makecell{Overall Err.\\(non-white) $\downarrow$} \\
    \midrule
    CAM16~\cite{li2017cam16} 
    & $804{,}803{,}899 \pm 17M$ 
    & $7.62 \pm 0.15\%$ 
    & $7.62 \pm 0.16\%$ 
    & $7.83 \pm 0.17\%$ \\

    CIE94~\cite{mcdonald1995cie94} 
    & $829{,}702{,}339 \pm 20M$ 
    & $7.85 \pm 0.18\%$ 
    & $7.85 \pm 0.19\%$ 
    & $8.07 \pm 0.20\%$ \\

    CIEDE2000~\cite{luo2001ciede2000} 
    & $795{,}231{,}886 \pm 14M$ 
    & $7.53 \pm 0.13\%$ 
    & $7.53 \pm 0.14\%$ 
    & $7.74 \pm 0.15\%$ \\

    CMC~\cite{clarke1984modification} 
    & $\textbf{768{,}638{,}180} \pm 12M$ 
    & $\textbf{7.27} \pm 0.10\%$ 
    & $\textbf{7.27} \pm 0.11\%$ 
    & $\textbf{7.48} \pm 0.12\%$ \\

    DIN99o~\cite{cui2002din99} 
    & $814{,}373{,}731 \pm 18M$ 
    & $7.71 \pm 0.16\%$ 
    & $7.71 \pm 0.17\%$ 
    & $7.93 \pm 0.18\%$ \\

    RGB 
    & $804{,}133{,}147 \pm 16M$ 
    & $7.61 \pm 0.14\%$ 
    & $7.61 \pm 0.15\%$ 
    & $7.83 \pm 0.16\%$ \\
    \bottomrule
  \end{tabular}
\end{table}

\subsection{Language Embedding Evaluation}
\label{sec:language_embedding_eval}

To evaluate language grounding in the final object-level map, we use 15 text queries and test whether the correct object is retrieved from the semantic Gaussian database. For each object, masked renderings are encoded with CLIP~\cite{radford2021learning} or Perception Encoder~\cite{bolya2025PerceptionEncoder}, using either hard masking, which keeps only the object, or soft masking, which retains limited context. As shown in Table~\ref{tab:embedding_bench}, soft masking improves retrieval across models, suggesting context aids recognition. Perception Encoder with soft masking performs best, achieving 80\% Top-1 and 87\% Top-3 accuracy. This supports storing one compact embedding per object instead of dense language features on every Gaussian.

\begin{table}[ht]
  \caption{Text-based object retrieval results on MonoVoc using 15 text queries.}
  \label{tab:embedding_bench}
  \centering
  \begin{tabular}{lccc}
    \toprule
    Embedding & Masking Type & Top-1 Accuracy (\%) $\uparrow$ & Top-3 Accuracy (\%) $\uparrow$ \\
    \midrule
    CLIP~\cite{radford2021learning} & Hard mask & $66.7 \pm 5.0$ & $73.3 \pm 3.6$ \\
    CLIP~\cite{radford2021learning} & Soft mask & $73.3 \pm 5.0$ & $80.0 \pm 6.2$ \\
    Perception Encoder~\cite{bolya2025PerceptionEncoder} & Hard mask & $73.3 \pm 7.1$ & $80.0 \pm 5.0$ \\
    Perception Encoder~\cite{bolya2025PerceptionEncoder} & Soft mask & $\mathbf{80.0 \pm 5.0}$ & $\mathbf{86.7 \pm 3.6}$ \\
    \bottomrule
  \end{tabular}
\end{table}

\subsection{Efficiency Analysis}
\label{sec:efficiency_analysis}

We also evaluate the efficiency of the generated 3D semantic maps in terms of map size, memory usage, and end-to-end runtime. As shown in Table~\ref{tab:efficiency_results}, MonoVoc produces a much smaller representation than the compared methods, using only 140K Gaussians per scene, versus 1M in ObjectGS~\cite{zhu2025objectgs} and 1.5M in SceneSplat~\cite{li2025scenesplat}. This compact map reduces memory usage to 14 MB per scene, nearly an order of magnitude below both baselines. Runtime is also considered: although SceneSplat reports a shorter time, it excludes 3D map generation, making the comparison less direct. In contrast, MonoVoc includes both SLAM-based reconstruction and semantic post-processing, better reflecting the cost of a full monocular pipeline. Overall, these results show that decoupling semantics from reconstruction and storing compact object-level representations, rather than dense per-Gaussian language features, yields a much lighter semantic map while remaining practical in memory and runtime.

\begin{table}[ht]
  \caption{Efficiency comparison in terms of average map size, average memory usage, and average runtime. MonoVoc uses fewer Gaussians and substantially less memory while maintaining a practical end-to-end runtime.}
  \label{tab:efficiency_results}
  \centering
  \begin{tabular}{lccc}
    \toprule
    Model & \# Gaussians per Scene $\downarrow$ & Memory Usage per Scene $\downarrow$ & Runtime $\downarrow$ \\
    \midrule
    ObjectGS~\cite{zhu2025objectgs} & 1M & 90 MB & 186 min $\pm$ 34 min\\
    SceneSplat~\cite{li2025scenesplat} & 1.5M & 149 MB & 40 min$^{*}$ $\pm$ 22 min\\
    MonoVoc (ours) & \textbf{140K} & \textbf{14 MB} & 81 min $\pm$ 17 min \\
    \bottomrule
  \end{tabular}
  \vspace{0.5em}
  
  \small{$^{*}$Runtime is reported without 3D map generation; SceneSplat does not generate the 3D map.}
\end{table}

\section{Discussion and Limitations}
\label{sec:discussion_limitations}

\paragraph{Discussion.}

The results support our main design choice: geometry and semantics need not be learned jointly for monocular open-vocabulary 3D understanding. Using HI-SLAM2~\cite{zhang2024hislam2} as a fixed geometric backbone, we add semantics after reconstruction without retraining the mapping pipeline, achieving strong rendering quality, competitive rendered segmentation accuracy, and a far more compact semantic map than ObjectGS and SceneSplat. Ablations show that deblending and semantic color correction are important, with CMC in Lab space~\cite{standard2007colorimetry,clarke1984modification} yielding the lowest semantic mismatch after alpha blending. For language grounding, Perception Encoder with soft masking performs best, though the pipeline is model- and segmenter-agnostic. Overall, compact object-level semantics suffice for efficient open-vocabulary retrieval and question answering without storing dense language features on every Gaussian.


\paragraph{Limitations.}

The method has two main limitations, both in semantic assignment. First, palette quantization can be ambiguous when object colors are close in the semantic palette or when deblending yields intermediate colors. We partly address this with CMC distance in Lab space, which is more perceptually meaningful than RGB distance, but performance could improve with better-separated or adaptive palettes. Second, SCD uses an efficient per-Gaussian approximation rather than solving the full global inverse problem over all overlapping Gaussians. This keeps the method lightweight, but may cause errors near object boundaries or in highly overlapping regions. Quantization reduces many mixed-color errors, while future work could add confidence weighting or sparse global refinement. Broader impact, privacy considerations, and deployment risks are discussed in Appendix~\ref{app:broader_impact}.



\section{Conclusion}
\label{sec:conclusion}



We presented a training-free pipeline for open-vocabulary 3D scene understanding from monocular videos. Rather than learning dense language features during reconstruction, our method first builds a geometry-aware Gaussian map with HI-SLAM2, then assigns semantics in post-processing. Semantic colors are recovered for individual Gaussians via deblending, refined through perceptual palette quantization, and linked to compact object-level language embeddings. This keeps the map readable and lightweight while supporting text-based object retrieval and question answering.

Experiments on Replica show that this modular design is effective, achieving strong reconstruction quality, competitive rendered segmentation accuracy, and substantially lower memory usage than baselines. Overall, our work suggests a simple direction for semantic 3D Gaussian maps: keep geometry, object identity, and language meaning modular. Although the method still depends on segmentation quality, color quantization, and approximate deblending, it provides a compact foundation for monocular open-vocabulary scene understanding. Future work can extend it to real-world videos, diverse environments, stronger object-level reasoning, and more robust handling of overlap, occlusion, and noisy semantic observations.

\bibliographystyle{plainnat}
\bibliography{sample-base}








\newpage
\appendix

\section{Additional deblending proofs}
\label{app:deblending_proofs}

This appendix gives the full derivations for the deblending formulation. We use
$v$ for a pixel coordinate and $m$ for a general observation, e.g., a pixel or a
pixel--frame pair. All semantic colors lie in $\mathbb{R}^3$, and all vector
equations hold channel-wise.

\paragraph{Lemma A.1 (Semantic-independence of alpha).}
\emph{Statement.}
Fix a pixel $v$. Let $i\in\{1,\dots,N\}$ index Gaussians ordered along the ray
through $v$. Let the unknown semantic color of Gaussian $i$ be
$s_i\in\mathbb{R}^3$. Let $\theta_i$ denote all non-semantic parameters of
Gaussian $i$, such as mean, covariance, and opacity. Define
\begin{equation}
\alpha_i(v)=g_i(v;\theta_i)\in[0,1),
\label{eq:app_alpha_indep_def}
\end{equation}
and assume $g_i$ does not depend on $\{s_k\}_{k=1}^{N}$. Then, for all $i,k$,
\begin{equation}
\frac{\partial \alpha_i(v)}{\partial s_k}=0.
\label{eq:app_alpha_indep_result}
\end{equation}

\emph{Proof.}
By assumption, $\alpha_i(v)=g_i(v;\theta_i)$ and $\theta_i$ is independent of
the semantic colors. Hence, for any $k$,
\begin{equation}
\frac{\partial \alpha_i(v)}{\partial s_k}
=
\frac{\partial}{\partial s_k}g_i(v;\theta_i)
=
0.
\end{equation}

\paragraph{Lemma A.2 (Semantic-independence of transmittance and weights).}
\emph{Statement.}
Assume Lemma A.1. Define
\begin{equation}
T_1(v)=1,
\label{eq:app_T1_def}
\end{equation}
\begin{equation}
T_i(v)=\prod_{j=1}^{i-1}\left(1-\alpha_j(v)\right)
\qquad
\mathrm{for}\quad i\ge 2,
\label{eq:app_Ti_def}
\end{equation}
\begin{equation}
w_i(v)=\alpha_i(v)T_i(v),
\label{eq:app_wi_def}
\end{equation}
and
\begin{equation}
T_{N+1}(v)=\prod_{j=1}^{N}\left(1-\alpha_j(v)\right).
\label{eq:app_TNp1_def}
\end{equation}
Then, for all $i,k$,
\begin{equation}
\frac{\partial T_i(v)}{\partial s_k}=0,
\qquad
\frac{\partial w_i(v)}{\partial s_k}=0,
\qquad
\frac{\partial T_{N+1}(v)}{\partial s_k}=0.
\label{eq:app_T_w_indep_result}
\end{equation}

\emph{Proof.}
For $i\ge 2$,
\begin{equation}
T_i(v)=\prod_{j=1}^{i-1}\left(1-\alpha_j(v)\right).
\end{equation}
Differentiating with respect to $s_k$ gives
\begin{equation}
\frac{\partial T_i(v)}{\partial s_k}
=
\sum_{\ell=1}^{i-1}
\left(
\prod_{\scriptstyle j=1,\ j\ne \ell}^{i-1}
\left(1-\alpha_j(v)\right)
\right)
\left(
-\frac{\partial \alpha_{\ell}(v)}{\partial s_k}
\right).
\end{equation}
By Lemma A.1, each derivative
$\partial \alpha_{\ell}(v)/\partial s_k$ is zero. Therefore
\begin{equation}
\frac{\partial T_i(v)}{\partial s_k}=0.
\end{equation}
Also $T_1(v)=1$, so $\partial T_1(v)/\partial s_k=0$. Since
$w_i(v)=\alpha_i(v)T_i(v)$,
\begin{equation}
\frac{\partial w_i(v)}{\partial s_k}
=
\frac{\partial \alpha_i(v)}{\partial s_k}T_i(v)
+
\alpha_i(v)\frac{\partial T_i(v)}{\partial s_k}
=
0.
\end{equation}
Finally, $T_{N+1}(v)$ is also a product of terms that are independent of the
semantic colors, so the same product-rule argument gives
\begin{equation}
\frac{\partial T_{N+1}(v)}{\partial s_k}=0.
\end{equation}

\paragraph{Theorem A.3 (Forward rendering is affine in semantic colors).}
\emph{Statement.}
Fix a pixel $v$. Let $s_i\in\mathbb{R}^3$ be unknown semantic colors and let
$bg\in\mathbb{R}^3$ be a known background color. The forward alpha-compositing
equation is
\begin{equation}
C(v)=\sum_{i=1}^{N}w_i(v)s_i+T_{N+1}(v)bg.
\label{eq:app_forward_affine}
\end{equation}
Then the map $(s_1,\dots,s_N)\mapsto C(v)$ is affine.

\emph{Proof.}
Let
\begin{equation}
S=
\left[
\begin{array}{c}
s_1\\
\vdots\\
s_N
\end{array}
\right]
\in\mathbb{R}^{3N}.
\label{eq:app_stacked_S}
\end{equation}
Define
\begin{equation}
A(v)=
\left[
\begin{array}{cccc}
w_1(v)I_3 & w_2(v)I_3 & \cdots & w_N(v)I_3
\end{array}
\right],
\label{eq:app_Av_def}
\end{equation}
and
\begin{equation}
b(v)=T_{N+1}(v)bg.
\label{eq:app_bv_def}
\end{equation}
Then
\begin{equation}
\sum_{i=1}^{N}w_i(v)s_i=A(v)S,
\end{equation}
and therefore
\begin{equation}
C(v)=A(v)S+b(v).
\end{equation}
By Lemma A.2, $A(v)$ and $b(v)$ are independent of the unknown semantic colors.
Thus the map from the semantic colors to $C(v)$ is affine.

\paragraph{Corollary A.4 (Linearity after background subtraction).}
\emph{Statement.}
Define the background-subtracted color
\begin{equation}
\widetilde{C}(v)=C(v)-T_{N+1}(v)bg.
\label{eq:app_C_tilde_def}
\end{equation}
Then
\begin{equation}
\widetilde{C}(v)=\sum_{i=1}^{N}w_i(v)s_i.
\label{eq:app_linear_after_bg}
\end{equation}

\emph{Proof.}
Using Equation~(\ref{eq:app_forward_affine}),
\begin{equation}
\widetilde{C}(v)
=
\left(
\sum_{i=1}^{N}w_i(v)s_i+T_{N+1}(v)bg
\right)
-
T_{N+1}(v)bg.
\end{equation}
The background terms cancel, giving
\begin{equation}
\widetilde{C}(v)=\sum_{i=1}^{N}w_i(v)s_i.
\end{equation}
Thus, after background subtraction, the inverse problem is linear in the unknown
semantic colors.

\paragraph{Corollary A.5 (Stacked linear system over many observations).}
\emph{Statement.}
Let $\mathcal{I}$ index observations, such as pixels or pixel--frame pairs. For
each $m\in\mathcal{I}$, define $w_{m,i}$ as the weight of Gaussian $i$ at
observation $m$ and $t_m=T_{N+1,m}$. Stack the observations as
\begin{equation}
Y=
\left[
\begin{array}{c}
C_{m_1}\\
\vdots\\
C_{m_{|\mathcal{I}|}}
\end{array}
\right],
\qquad
B=
\left[
\begin{array}{c}
t_{m_1}bg\\
\vdots\\
t_{m_{|\mathcal{I}|}}bg
\end{array}
\right].
\label{eq:app_stack_YB}
\end{equation}
Then there exists a matrix $W$ with block rows
\begin{equation}
W_m=
\left[
\begin{array}{cccc}
w_{m,1}I_3 & w_{m,2}I_3 & \cdots & w_{m,N}I_3
\end{array}
\right]
\label{eq:app_stack_W_row}
\end{equation}
such that
\begin{equation}
Y=WS+B,
\qquad
Y-B=WS.
\label{eq:app_stacked_linear_system}
\end{equation}

\emph{Proof.}
For each observation $m\in\mathcal{I}$, Theorem A.3 gives
\begin{equation}
C_m=\sum_{i=1}^{N}w_{m,i}s_i+t_m bg.
\end{equation}
Stacking these equations over all observations gives
\begin{equation}
Y=WS+B.
\end{equation}
Subtracting $B$ from both sides gives
\begin{equation}
Y-B=WS.
\end{equation}

\paragraph{Lemma A.6 (Recursive transmittance update).}
\emph{Statement.}
Fix a pixel $v$ and define
\begin{equation}
T_1(v)=1,
\qquad
T_i(v)=\prod_{j=1}^{i-1}\left(1-\alpha_j(v)\right)
\quad
\mathrm{for}\quad i\ge 2.
\label{eq:app_recursive_T_def}
\end{equation}
Then, for all $i\in\{1,\dots,N\}$,
\begin{equation}
T_{i+1}(v)=T_i(v)\left(1-\alpha_i(v)\right).
\label{eq:app_recursive_T_result}
\end{equation}

\emph{Proof.}
By definition,
\begin{equation}
T_{i+1}(v)
=
\prod_{j=1}^{i}\left(1-\alpha_j(v)\right)
=
\left(
\prod_{j=1}^{i-1}\left(1-\alpha_j(v)\right)
\right)
\left(1-\alpha_i(v)\right).
\end{equation}
The product in parentheses is $T_i(v)$, so
\begin{equation}
T_{i+1}(v)=T_i(v)\left(1-\alpha_i(v)\right).
\end{equation}

\paragraph{Lemma A.7 (Weight equals transmittance drop).}
\emph{Statement.}
Let
\begin{equation}
w_i(v)=\alpha_i(v)T_i(v).
\label{eq:app_weight_drop_w_def}
\end{equation}
Then, for all $i\in\{1,\dots,N\}$,
\begin{equation}
w_i(v)=T_i(v)-T_{i+1}(v).
\label{eq:app_weight_drop_result}
\end{equation}

\emph{Proof.}
Using Lemma A.6,
\begin{equation}
T_i(v)-T_{i+1}(v)
=
T_i(v)-T_i(v)\left(1-\alpha_i(v)\right).
\end{equation}
Therefore,
\begin{equation}
T_i(v)-T_{i+1}(v)
=
T_i(v)\alpha_i(v)
=
w_i(v).
\end{equation}

\paragraph{Theorem A.8 (Partition of unity).}
\emph{Statement.}
Let
\begin{equation}
T_{N+1}(v)=\prod_{j=1}^{N}\left(1-\alpha_j(v)\right).
\label{eq:app_partition_TNp1}
\end{equation}
Then
\begin{equation}
\sum_{i=1}^{N}w_i(v)=1-T_{N+1}(v),
\qquad
\sum_{i=1}^{N}w_i(v)+T_{N+1}(v)=1.
\label{eq:app_partition_result}
\end{equation}

\emph{Proof.}
By Lemma A.7,
\begin{equation}
\sum_{i=1}^{N}w_i(v)
=
\sum_{i=1}^{N}\left(T_i(v)-T_{i+1}(v)\right).
\end{equation}
This sum telescopes:
\begin{equation}
\sum_{i=1}^{N}w_i(v)
=
\left(T_1(v)-T_2(v)\right)
+
\cdots
+
\left(T_N(v)-T_{N+1}(v)\right)
=
T_1(v)-T_{N+1}(v).
\end{equation}
Since $T_1(v)=1$,
\begin{equation}
\sum_{i=1}^{N}w_i(v)=1-T_{N+1}(v).
\end{equation}
Adding $T_{N+1}(v)$ to both sides gives
\begin{equation}
\sum_{i=1}^{N}w_i(v)+T_{N+1}(v)=1.
\end{equation}

\paragraph{Corollary A.9 (Nonnegativity and boundedness).}
\emph{Statement.}
If $\alpha_i(v)\in[0,1)$, then, for all $i$,
\begin{equation}
0\le T_{i+1}(v)\le T_i(v)\le 1,
\qquad
0\le w_i(v)\le 1,
\qquad
0\le T_{N+1}(v)\le 1.
\label{eq:app_nonnegative_bounds}
\end{equation}

\emph{Proof.}
Since $\alpha_i(v)\in[0,1)$, each factor
$1-\alpha_i(v)$ lies in $(0,1]$. Therefore every transmittance product is
nonnegative and at most one:
\begin{equation}
0\le T_i(v)\le 1.
\end{equation}
By Lemma A.6,
\begin{equation}
T_{i+1}(v)=T_i(v)\left(1-\alpha_i(v)\right),
\end{equation}
so $0\le T_{i+1}(v)\le T_i(v)$. By Lemma A.7,
\begin{equation}
w_i(v)=T_i(v)-T_{i+1}(v).
\end{equation}
Thus $w_i(v)\ge 0$ and $w_i(v)\le T_i(v)\le 1$. The same product argument gives
$0\le T_{N+1}(v)\le 1$.

\paragraph{Theorem A.10 (Exact mixing form of alpha compositing).}
\emph{Statement.}
The rendered color
\begin{equation}
C(v)=\sum_{i=1}^{N}w_i(v)s_i+T_{N+1}(v)bg
\label{eq:app_exact_mixing}
\end{equation}
is an exact convex mixture of the Gaussian semantic colors and the background:
\begin{equation}
w_i(v)\ge 0,
\qquad
T_{N+1}(v)\ge 0,
\qquad
\sum_{i=1}^{N}w_i(v)+T_{N+1}(v)=1.
\label{eq:app_exact_mixing_conditions}
\end{equation}

\emph{Proof.}
The nonnegativity of $w_i(v)$ and $T_{N+1}(v)$ follows from Corollary A.9. The
sum-to-one property follows from Theorem A.8. Therefore the coefficients in
Equation~(\ref{eq:app_exact_mixing}) form a convex mixture over the semantic
colors $\{s_i\}_{i=1}^{N}$ and the background $bg$.

\paragraph{Lemma A.11 (Convex quadratic form).}
\emph{Statement.}
Fix a Gaussian index $i$. Let $\mathcal{M}$ index observations. For each
$m\in\mathcal{M}$, let $y_m\in\mathbb{R}^3$ be a known target and
$w_{i,m}\ge 0$ be the known contribution weight of Gaussian $i$. Define
\begin{equation}
J_i(s)=\sum_{m\in\mathcal{M}}w_{i,m}\left\|s-y_m\right\|_2^2.
\label{eq:app_quad_objective}
\end{equation}
Then $J_i$ is a convex quadratic in $s$. Its Hessian is
\begin{equation}
\nabla^2 J_i(s)=2\left(\sum_{m\in\mathcal{M}}w_{i,m}\right)I_3.
\label{eq:app_quad_hessian}
\end{equation}
If $\sum_{m\in\mathcal{M}}w_{i,m}>0$, then $J_i$ is strictly convex and has a
unique minimizer.

\emph{Proof.}
We expand the squared norm:
\begin{equation}
\left\|s-y_m\right\|_2^2=(s-y_m)^{\top}(s-y_m).
\end{equation}
Hence
\begin{equation}
J_i(s)=\sum_m w_{i,m}(s-y_m)^{\top}(s-y_m).
\end{equation}
Differentiating gives
\begin{equation}
\nabla J_i(s)
=
\sum_m 2w_{i,m}(s-y_m)
=
2\left(\sum_m w_{i,m}\right)s
-
2\sum_m w_{i,m}y_m.
\label{eq:app_quad_gradient}
\end{equation}
Differentiating again gives
\begin{equation}
\nabla^2 J_i(s)=2\left(\sum_m w_{i,m}\right)I_3.
\end{equation}
If $\sum_m w_{i,m}>0$, the Hessian is positive definite, so $J_i$ is strictly
convex and the minimizer is unique.

\paragraph{Theorem A.12 (Weighted-average estimator).}
\emph{Statement.}
Assume
\begin{equation}
W_i=\sum_{m\in\mathcal{M}}w_{i,m}>0.
\label{eq:app_Wi_def}
\end{equation}
Then the unique minimizer of
\begin{equation}
\hat{s}_i
=
{\rm arg\,min}_{s\in\mathbb{R}^3}
\sum_{m\in\mathcal{M}}w_{i,m}\left\|s-y_m\right\|_2^2
\label{eq:app_weighted_average_problem}
\end{equation}
is
\begin{equation}
\hat{s}_i=
\frac{\sum_{m\in\mathcal{M}}w_{i,m}y_m}
{\sum_{m\in\mathcal{M}}w_{i,m}}.
\label{eq:app_weighted_average_solution}
\end{equation}

\emph{Proof.}
By Lemma A.11, the objective is strictly convex when $W_i>0$, so the unique
minimizer satisfies $\nabla J_i(s)=0$. From Equation~(\ref{eq:app_quad_gradient}),
\begin{equation}
0=
2W_i s
-
2\sum_m w_{i,m}y_m.
\end{equation}
Thus
\begin{equation}
W_i s=\sum_m w_{i,m}y_m.
\end{equation}
Since $W_i>0$,
\begin{equation}
s=
\frac{\sum_m w_{i,m}y_m}{W_i}.
\end{equation}
Therefore the minimizer is Equation~(\ref{eq:app_weighted_average_solution}).

\paragraph{Corollary A.13 (Accumulation form).}
\emph{Statement.}
Define
\begin{equation}
\mathrm{color\_accum}[i]
=
\sum_{m\in\mathcal{M}}w_{i,m}y_m,
\qquad
\mathrm{weight\_accum}[i]
=
\sum_{m\in\mathcal{M}}w_{i,m}.
\label{eq:app_accum_def}
\end{equation}
If $\mathrm{weight\_accum}[i]>0$, then
\begin{equation}
\hat{s}_i=
\frac{\mathrm{color\_accum}[i]}
{\mathrm{weight\_accum}[i]}.
\label{eq:app_accum_solution}
\end{equation}

\emph{Proof.}
This follows directly from Theorem A.12 by substituting the two accumulator
definitions from Equation~(\ref{eq:app_accum_def}).

\paragraph{Theorem A.14 (Exact recovery under one-hot mixing).}
\emph{Statement.}
Let $\mathcal{M}$ index observations. Assume the forward model
\begin{equation}
C_m=\sum_{i=1}^{N}w_{i,m}s_i+T_{N+1,m}bg
\label{eq:app_onehot_forward}
\end{equation}
and define
\begin{equation}
y_m=C_m-T_{N+1,m}bg.
\label{eq:app_onehot_y}
\end{equation}
Assume one-hot mixing: for every $m\in\mathcal{M}$, there exists
$\iota(m)\in\{1,\dots,N\}$ such that
\begin{equation}
w_{\iota(m),m}=1,
\qquad
w_{j,m}=0
\quad
\mathrm{for}\quad j\ne \iota(m).
\label{eq:app_onehot_condition}
\end{equation}
Let
\begin{equation}
\mathcal{M}_i=\{m\in\mathcal{M}:\iota(m)=i\}.
\label{eq:app_Mi_def}
\end{equation}
If $|\mathcal{M}_i|>0$, then the estimator
\begin{equation}
\hat{s}_i=
\frac{\sum_{m\in\mathcal{M}}w_{i,m}y_m}
{\sum_{m\in\mathcal{M}}w_{i,m}}
\label{eq:app_onehot_estimator}
\end{equation}
recovers the true semantic color:
\begin{equation}
\hat{s}_i=s_i.
\label{eq:app_onehot_result}
\end{equation}

\emph{Proof.}
From Equations~(\ref{eq:app_onehot_forward}) and
(\ref{eq:app_onehot_y}),
\begin{equation}
y_m=\sum_{i=1}^{N}w_{i,m}s_i.
\end{equation}
Under one-hot mixing,
\begin{equation}
y_m
=
w_{\iota(m),m}s_{\iota(m)}
+
\sum_{j\ne\iota(m)}w_{j,m}s_j
=
s_{\iota(m)}.
\end{equation}
Fix a Gaussian $i$ with $|\mathcal{M}_i|>0$. Then
\begin{equation}
\sum_{m\in\mathcal{M}}w_{i,m}y_m
=
\sum_{m\in\mathcal{M}_i}y_m
=
\sum_{m\in\mathcal{M}_i}s_i
=
|\mathcal{M}_i|s_i.
\end{equation}
Also,
\begin{equation}
\sum_{m\in\mathcal{M}}w_{i,m}
=
\sum_{m\in\mathcal{M}_i}1
=
|\mathcal{M}_i|.
\end{equation}
Therefore
\begin{equation}
\hat{s}_i
=
\frac{|\mathcal{M}_i|s_i}{|\mathcal{M}_i|}
=
s_i.
\end{equation}

\paragraph{Theorem A.15 (Overlap decomposition of the estimator).}
\emph{Statement.}
Assume the background-subtracted forward model
\begin{equation}
y_m=\sum_{j=1}^{N}w_{j,m}s_j.
\label{eq:app_overlap_forward}
\end{equation}
Define
\begin{equation}
W_i=\sum_{m\in\mathcal{M}}w_{i,m},
\qquad
G_{ij}=\sum_{m\in\mathcal{M}}w_{i,m}w_{j,m}.
\label{eq:app_overlap_W_G}
\end{equation}
If $W_i>0$, then the estimator
\begin{equation}
\hat{s}_i=
\frac{\sum_{m\in\mathcal{M}}w_{i,m}y_m}{W_i}
\label{eq:app_overlap_estimator}
\end{equation}
satisfies
\begin{equation}
\hat{s}_i
=
\frac{G_{ii}}{W_i}s_i
+
\sum_{j=1,\,j\ne i}^{N}
\frac{G_{ij}}{W_i}s_j.
\label{eq:app_overlap_decomp}
\end{equation}

\emph{Proof.}
Substitute Equation~(\ref{eq:app_overlap_forward}) into
Equation~(\ref{eq:app_overlap_estimator}):
\begin{equation}
\hat{s}_i
=
\frac{1}{W_i}
\sum_{m\in\mathcal{M}}w_{i,m}
\left(
\sum_{j=1}^{N}w_{j,m}s_j
\right).
\end{equation}
Rearranging the sums gives
\begin{equation}
\hat{s}_i
=
\frac{1}{W_i}
\sum_{j=1}^{N}
\left(
\sum_{m\in\mathcal{M}}w_{i,m}w_{j,m}
\right)s_j.
\end{equation}
Using the definition of $G_{ij}$,
\begin{equation}
\hat{s}_i
=
\frac{1}{W_i}
\sum_{j=1}^{N}G_{ij}s_j.
\end{equation}
Separating the $j=i$ term gives Equation~(\ref{eq:app_overlap_decomp}).

\paragraph{Corollary A.16 (Explicit bias under overlap).}
\emph{Statement.}
Define
\begin{equation}
e_i=\hat{s}_i-s_i.
\label{eq:app_error_def}
\end{equation}
Then
\begin{equation}
e_i
=
\left(\frac{G_{ii}}{W_i}-1\right)s_i
+
\sum_{j=1,\,j\ne i}^{N}
\frac{G_{ij}}{W_i}s_j.
\label{eq:app_error_bias_1}
\end{equation}
Equivalently,
\begin{equation}
e_i
=
-\frac{W_i-G_{ii}}{W_i}s_i
+
\sum_{j=1,\,j\ne i}^{N}
\frac{G_{ij}}{W_i}s_j,
\label{eq:app_error_bias_2}
\end{equation}
where
\begin{equation}
W_i-G_{ii}
=
\sum_{m\in\mathcal{M}}w_{i,m}(1-w_{i,m}).
\label{eq:app_W_minus_G}
\end{equation}

\emph{Proof.}
Subtract $s_i$ from Equation~(\ref{eq:app_overlap_decomp}):
\begin{equation}
e_i
=
\hat{s}_i-s_i
=
\left(\frac{G_{ii}}{W_i}-1\right)s_i
+
\sum_{j=1,\,j\ne i}^{N}
\frac{G_{ij}}{W_i}s_j.
\end{equation}
Also,
\begin{equation}
W_i-G_{ii}
=
\sum_m w_{i,m}
-
\sum_m w_{i,m}^2
=
\sum_m w_{i,m}(1-w_{i,m}).
\end{equation}
Therefore
\begin{equation}
\frac{G_{ii}}{W_i}-1
=
-\frac{W_i-G_{ii}}{W_i},
\end{equation}
which gives Equation~(\ref{eq:app_error_bias_2}).

\paragraph{Corollary A.17 (Norm bound in terms of overlap).}
\emph{Statement.}
Under Corollary A.16,
\begin{equation}
\left\|e_i\right\|_2
\le
\left|\frac{G_{ii}}{W_i}-1\right|
\left\|s_i\right\|_2
+
\sum_{j=1,\,j\ne i}^{N}
\frac{G_{ij}}{W_i}
\left\|s_j\right\|_2.
\label{eq:app_overlap_bound}
\end{equation}

\emph{Proof.}
Apply the triangle inequality to Equation~(\ref{eq:app_error_bias_1}):
\begin{equation}
\left\|e_i\right\|_2
=
\left\|
\left(\frac{G_{ii}}{W_i}-1\right)s_i
+
\sum_{j=1,\,j\ne i}^{N}
\frac{G_{ij}}{W_i}s_j
\right\|_2.
\end{equation}
Thus,
\begin{equation}
\left\|e_i\right\|_2
\le
\left|\frac{G_{ii}}{W_i}-1\right|
\left\|s_i\right\|_2
+
\sum_{j=1,\,j\ne i}^{N}
\frac{G_{ij}}{W_i}
\left\|s_j\right\|_2.
\end{equation}

\paragraph{Lemma A.18 (Normal equations of the global least-squares inverse).}
\emph{Statement.}
Let $\mathcal{M}$ index $M$ observations. Stack the unknown semantic colors as
\begin{equation}
S=
\left[
\begin{array}{c}
s_1^{\top}\\
\vdots\\
s_N^{\top}
\end{array}
\right]
\in\mathbb{R}^{N\times 3},
\label{eq:app_global_S}
\end{equation}
and stack the background-subtracted targets as
\begin{equation}
Y=
\left[
\begin{array}{c}
y_{m_1}^{\top}\\
\vdots\\
y_{m_M}^{\top}
\end{array}
\right]
\in\mathbb{R}^{M\times 3}.
\label{eq:app_global_Y}
\end{equation}
Let $W\in\mathbb{R}^{M\times N}$ be the matrix with entries
\begin{equation}
W_{m,i}=w_{m,i}.
\label{eq:app_global_W}
\end{equation}
Consider
\begin{equation}
S^{\star}
=
{\rm arg\,min}_{S\in\mathbb{R}^{N\times 3}}
\left\|WS-Y\right\|_F^2.
\label{eq:app_global_ls}
\end{equation}
Then $S^{\star}$ satisfies
\begin{equation}
(W^{\top}W)S=W^{\top}Y.
\label{eq:app_normal_equations}
\end{equation}
Equivalently, defining
\begin{equation}
G=W^{\top}W,
\qquad
B=W^{\top}Y,
\label{eq:app_GB_def}
\end{equation}
the system is
\begin{equation}
GS=B,
\label{eq:app_GS_B}
\end{equation}
with entries
\begin{equation}
G_{ij}=\sum_{m\in\mathcal{M}}w_{m,i}w_{m,j},
\qquad
B_i=\sum_{m\in\mathcal{M}}w_{m,i}y_m.
\label{eq:app_GB_entries}
\end{equation}

\emph{Proof.}
Expand the objective:
\begin{equation}
\left\|WS-Y\right\|_F^2
=
\mathrm{tr}\left((WS-Y)^{\top}(WS-Y)\right).
\end{equation}
Differentiating with respect to $S$ gives
\begin{equation}
\nabla_S \left\|WS-Y\right\|_F^2
=
2W^{\top}(WS-Y).
\end{equation}
At a minimizer, the gradient is zero:
\begin{equation}
W^{\top}(WS-Y)=0.
\end{equation}
Therefore
\begin{equation}
(W^{\top}W)S=W^{\top}Y.
\end{equation}
Using $G=W^{\top}W$ and $B=W^{\top}Y$ gives $GS=B$. Entrywise,
\begin{equation}
G_{ij}
=
(W^{\top}W)_{ij}
=
\sum_m W_{m,i}W_{m,j}
=
\sum_m w_{m,i}w_{m,j},
\end{equation}
and
\begin{equation}
B_i
=
(W^{\top}Y)_i
=
\sum_m W_{m,i}Y_m
=
\sum_m w_{m,i}y_m.
\end{equation}

\paragraph{Theorem A.19 (Approach 1 as a diagonal approximation).}
\emph{Statement.}
The exact normal equations can be written row-wise as
\begin{equation}
\sum_{j=1}^{N}G_{ij}s_j=B_i.
\label{eq:app_normal_row}
\end{equation}
Using the diagonal-only approximation
\begin{equation}
G\approx D=\mathrm{diag}(G_{11},\dots,G_{NN})
\label{eq:app_diag_approx}
\end{equation}
gives the decoupled estimator
\begin{equation}
\widetilde{s}_i
=
\frac{B_i}{G_{ii}}
=
\frac{\sum_{m\in\mathcal{M}}w_{m,i}y_m}
{\sum_{m\in\mathcal{M}}w_{m,i}^2},
\qquad
G_{ii}>0.
\label{eq:app_diag_estimator}
\end{equation}
If $w_{m,i}^2=w_{m,i}$ for all $m$, then $G_{ii}=W_i$ and
\begin{equation}
\widetilde{s}_i
=
\frac{\sum_{m\in\mathcal{M}}w_{m,i}y_m}
{\sum_{m\in\mathcal{M}}w_{m,i}}
=
\hat{s}_i.
\label{eq:app_diag_to_approach1}
\end{equation}
More generally, if $w_{m,i}^2\approx w_{m,i}$, then
$\widetilde{s}_i\approx \hat{s}_i$.

\emph{Proof.}
From Equation~(\ref{eq:app_normal_row}),
\begin{equation}
G_{ii}s_i+\sum_{j=1,\,j\ne i}^{N}G_{ij}s_j=B_i.
\end{equation}
The diagonal approximation sets the off-diagonal terms $G_{ij}$, $j\ne i$, to
zero, giving
\begin{equation}
G_{ii}s_i\approx B_i.
\end{equation}
Thus
\begin{equation}
s_i\approx \frac{B_i}{G_{ii}}=\widetilde{s}_i.
\end{equation}
If $w_{m,i}^2=w_{m,i}$ for all $m$, then
\begin{equation}
G_{ii}
=
\sum_m w_{m,i}^2
=
\sum_m w_{m,i}
=
W_i.
\end{equation}
Substituting $G_{ii}=W_i$ into Equation~(\ref{eq:app_diag_estimator}) gives
\begin{equation}
\widetilde{s}_i
=
\frac{\sum_m w_{m,i}y_m}{\sum_m w_{m,i}}
=
\hat{s}_i.
\end{equation}
If $w_{m,i}^2\approx w_{m,i}$, then
\begin{equation}
G_{ii}=\sum_m w_{m,i}^2\approx \sum_m w_{m,i}=W_i,
\end{equation}
so the diagonal estimator is close to the weighted-average estimator.

\paragraph{Theorem A.20 (Multi-frame weighted average as a single-index weighted LS).}
\emph{Statement.}
Let $\mathcal{F}$ index frames and let $\mathcal{P}_f$ index pixels in frame
$f$. Define
\begin{equation}
\mathcal{M}=\{(p,f): f\in\mathcal{F},\ p\in\mathcal{P}_f\}.
\label{eq:app_multiframe_M}
\end{equation}
For each observation $(p,f)$, let $y_{p,f}\in\mathbb{R}^3$ be a known target and
let $w_{i,p,f}\ge 0$ be the known weight of Gaussian $i$. If
\begin{equation}
W_i=\sum_{f\in\mathcal{F}}\sum_{p\in\mathcal{P}_f}w_{i,p,f}>0,
\label{eq:app_multiframe_Wi}
\end{equation}
then the minimizer of
\begin{equation}
\hat{s}_i
=
{\rm arg\,min}_{s\in\mathbb{R}^3}
\sum_{f\in\mathcal{F}}\sum_{p\in\mathcal{P}_f}
w_{i,p,f}\left\|s-y_{p,f}\right\|_2^2
\label{eq:app_multiframe_objective}
\end{equation}
is
\begin{equation}
\hat{s}_i
=
\frac{
\sum_{f\in\mathcal{F}}\sum_{p\in\mathcal{P}_f}
w_{i,p,f}y_{p,f}
}{
\sum_{f\in\mathcal{F}}\sum_{p\in\mathcal{P}_f}
w_{i,p,f}
}.
\label{eq:app_multiframe_solution}
\end{equation}

\emph{Proof.}
Using the observation set $\mathcal{M}$, the objective can be written as
\begin{equation}
\hat{s}_i
=
{\rm arg\,min}_{s\in\mathbb{R}^3}
\sum_{(p,f)\in\mathcal{M}}
w_{i,p,f}\left\|s-y_{p,f}\right\|_2^2.
\end{equation}
Differentiating and setting the derivative to zero gives
\begin{equation}
0
=
\sum_{(p,f)\in\mathcal{M}}2w_{i,p,f}(s-y_{p,f}).
\end{equation}
Therefore
\begin{equation}
0
=
2W_i s
-
2\sum_{(p,f)\in\mathcal{M}}w_{i,p,f}y_{p,f}.
\end{equation}
Thus
\begin{equation}
W_i s
=
\sum_{(p,f)\in\mathcal{M}}w_{i,p,f}y_{p,f}.
\end{equation}
Since $W_i>0$,
\begin{equation}
\hat{s}_i
=
\frac{\sum_{(p,f)\in\mathcal{M}}w_{i,p,f}y_{p,f}}{W_i}.
\end{equation}
Expanding $\mathcal{M}$ back into frames and pixels gives
Equation~(\ref{eq:app_multiframe_solution}).

\paragraph{Corollary A.21 (Accumulation across frames).}
\emph{Statement.}
Define
\begin{equation}
\mathrm{color\_accum}[i]
=
\sum_{f\in\mathcal{F}}\sum_{p\in\mathcal{P}_f}
w_{i,p,f}y_{p,f},
\qquad
\mathrm{weight\_accum}[i]
=
\sum_{f\in\mathcal{F}}\sum_{p\in\mathcal{P}_f}
w_{i,p,f}.
\label{eq:app_multiframe_accum_def}
\end{equation}
If $\mathrm{weight\_accum}[i]>0$, then
\begin{equation}
\hat{s}_i
=
\frac{\mathrm{color\_accum}[i]}
{\mathrm{weight\_accum}[i]}.
\label{eq:app_multiframe_accum_solution}
\end{equation}

\emph{Proof.}
Substituting the accumulator definitions in
Equation~(\ref{eq:app_multiframe_accum_def}) into
Equation~(\ref{eq:app_multiframe_solution}) gives
Equation~(\ref{eq:app_multiframe_accum_solution}).

\paragraph{Corollary A.22 (Stacked multi-frame linear-system interpretation).}
\emph{Statement.}
Let $S\in\mathbb{R}^{N\times 3}$ stack the Gaussian semantic colors, and let
$Y\in\mathbb{R}^{|\mathcal{M}|\times 3}$ stack the targets
$\{y_{p,f}\}$. Let $W\in\mathbb{R}^{|\mathcal{M}|\times N}$ have entries
\begin{equation}
W_{(p,f),i}=w_{i,p,f}.
\label{eq:app_multiframe_W_entries}
\end{equation}
Then the multi-frame forward model is
\begin{equation}
Y=WS,
\label{eq:app_multiframe_stacked}
\end{equation}
and the estimator in Theorem A.20 is the weighted least-squares minimizer over
the expanded index set $\mathcal{M}$.

\emph{Proof.}
By definition, each observation $(p,f)$ satisfies
\begin{equation}
y_{p,f}=\sum_{i=1}^{N}w_{i,p,f}s_i.
\end{equation}
Stacking all observations gives
\begin{equation}
Y=WS.
\end{equation}
The multi-frame objective in Equation~(\ref{eq:app_multiframe_objective}) is
therefore the same per-Gaussian weighted least-squares problem as before, but
with the observation index set expanded to include all pixels across all frames.

\section{Why Overlap Happens?}
\label{app:overlap}

\begin{figure}[h]
  \centering
  \includegraphics[width=\linewidth]{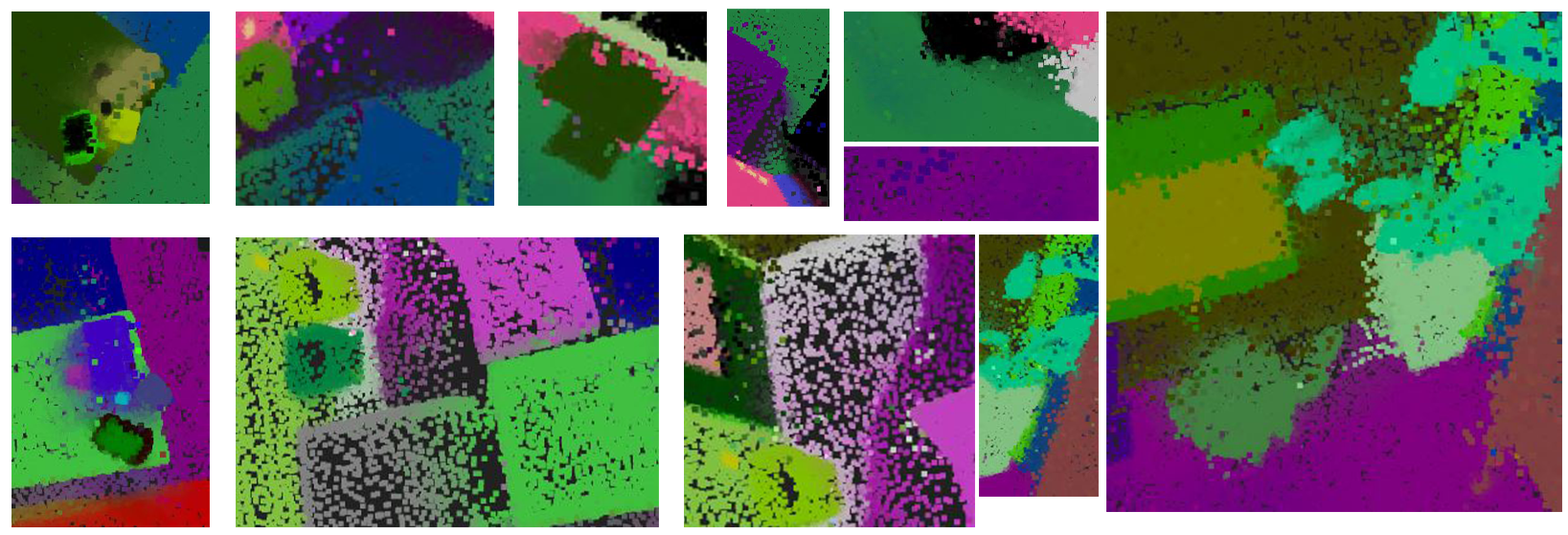}
  \caption{
  Typical overlap cases in semantic Gaussian deblending.
  }
  \label{fig:overlap_cases}
\end{figure}

This appendix explains why semantic overlap appears in our pipeline and why it is
a natural consequence of alpha-composited Gaussian rendering. The main point is
that a rendered pixel is rarely produced by a single Gaussian. Instead, it is
usually formed by several projected Gaussians whose contributions are blended in
image space. Our deblending step tries to reverse this process from the rendered
2D observations, but when the same Gaussian contributes to pixels belonging to
different semantic regions, its recovered color can become a mixture of those
regions. Figure~\ref{fig:overlap_cases} illustrates common situations where overlap
appears. 

\paragraph{Overlap is caused by image-space alpha blending.}
In Gaussian splatting, each 3D Gaussian is projected to a 2D elliptical footprint
on the image plane. This footprint is not a hard region. Its opacity is highest
near the projected center and decreases smoothly toward the boundary according to
the projected covariance. Therefore, even if a Gaussian mainly belongs to one
object, its projected footprint may still cover nearby pixels from another
object, especially near object boundaries.

\begin{figure}[h]
  \centering
  \includegraphics[width=0.75\linewidth,height=0.55\linewidth]{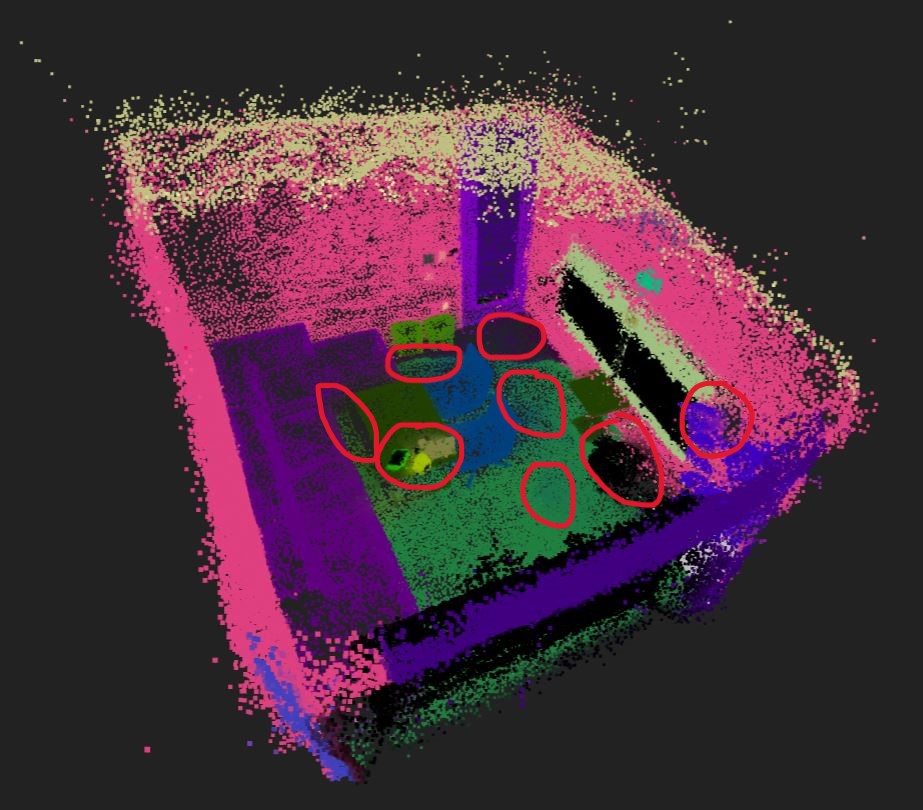}
  \caption{Visualization of the overlap effect in semantic deblending.}
  \label{fig:appendix_deblending_overlap}
\end{figure}

For a pixel observation $p$, the background-subtracted semantic color can be
written as
\begin{equation}
y_p = \sum_{i=1}^{N} w_{i,p}s_i,
\label{eq:appendix_overlap_forward}
\end{equation}
where $s_i$ is the semantic color of Gaussian $i$ and $w_{i,p}$ is its
alpha-compositing weight at pixel $p$. Equation~\ref{eq:appendix_overlap_forward}
shows that the observed color at a pixel is a weighted mixture of all Gaussians
that influence that pixel. If only one Gaussian contributes to the pixel, then
the observation directly represents that Gaussian. However, when several
Gaussians overlap in the same pixel, the observation contains semantic evidence
from multiple objects at once (Figure~\ref{fig:appendix_deblending_overlap}).

\paragraph{Object boundaries are the most affected regions.}
Overlap is especially visible around object boundaries. At the boundary between
two objects, the projected Gaussian footprints from both sides can cover the same
pixels. For example, a Gaussian located on the edge of a chair may still splat
partly onto pixels labeled as the wall or the floor. Similarly, a Gaussian on the
edge of a table may influence both table pixels and background pixels. Since the
renderer blends these contributions smoothly, the resulting semantic observation
near the boundary may lie between the valid object colors.

This does not mean that the geometry is incorrect. It is a normal result of using
soft Gaussian primitives rather than hard object surfaces. A Gaussian is a smooth
basis function with spatial extent, so its influence naturally spreads beyond a
single ideal pixel or object mask.

\paragraph{The inverse problem is coupled.}
Our deblending step assigns a semantic color to each Gaussian by collecting all
pixel observations in which that Gaussian participates:
\begin{equation}
\hat{s}_i =
\frac{\sum_p w_{i,p}y_p}{\sum_p w_{i,p}}.
\label{eq:appendix_overlap_average}
\end{equation}
This estimator is exact when the Gaussian only contributes to pixels of one
semantic color. However, if Gaussian $i$ also contributes to pixels where other
Gaussians are visible, then the observations $y_p$ already contain the colors of
those other Gaussians. Substituting
Equation~\ref{eq:appendix_overlap_forward} into
Equation~\ref{eq:appendix_overlap_average} gives
\begin{equation}
\hat{s}_i =
\frac{1}{W_i}
\sum_{j=1}^{N}
\left(\sum_p w_{i,p}w_{j,p}\right)s_j,
\qquad
W_i=\sum_p w_{i,p}.
\label{eq:appendix_overlap_bias}
\end{equation}
The term
\begin{equation}
G_{ij}=\sum_p w_{i,p}w_{j,p}
\end{equation}
measures how often Gaussian $i$ and Gaussian $j$ influence the same observations.
When $G_{ij}$ is large for $j\neq i$, the recovered color of Gaussian $i$ is
pulled toward the color of Gaussian $j$. This is the overlap bias described in
Theorem~4. In other words, the deblending formula is solving the best
per-Gaussian average, but the rendered observations themselves are not purely
per-Gaussian; they are already mixtures produced by the forward renderer.

\paragraph{Multi-view observations can increase ambiguity.}
The same Gaussian is seen from many frames and viewing directions. In one frame,
it may project mostly inside its own object mask. In another frame, due to a
different viewpoint, camera pose, occlusion ordering, or segmentation boundary,
the same Gaussian may partially overlap a neighboring object. Therefore, the
semantic evidence collected for one Gaussian may not be perfectly consistent
across all frames.

This is important because our estimator uses all observations where the Gaussian
has nonzero weight. A Gaussian that is mostly assigned to one object can still
receive weaker evidence from another object if it repeatedly overlaps that object
in some views. As a result, its averaged semantic color may drift away from the
true palette color and move toward an intermediate value.

\paragraph{Small opacity does not always mean small accumulated influence.}
Although a Gaussian may have only a small contribution to a single wrong pixel,
this effect can accumulate across many frames. If the Gaussian appears in many
views and repeatedly touches a neighboring semantic region, the total weighted
evidence from that region can become noticeable. Thus, overlap is not only a
single-frame boundary artifact; it is also a multi-view accumulation effect.

\begin{figure}[h]
  \centering
  \includegraphics[width=0.85\linewidth]{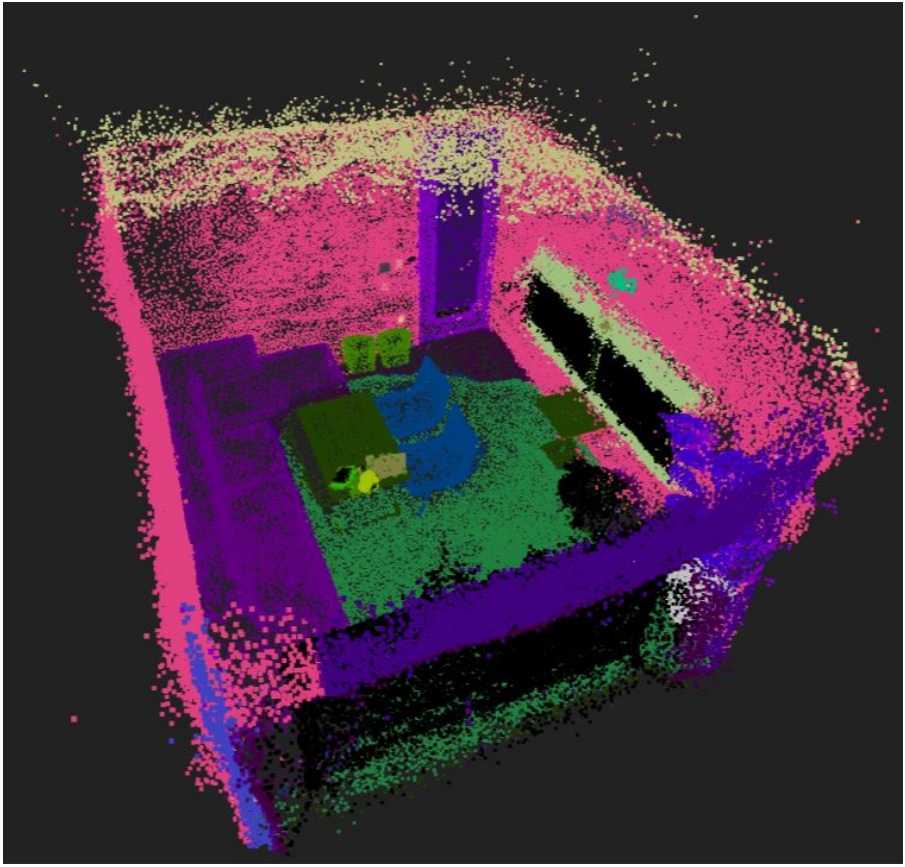}
  \caption{Visualization of CMC-based semantic palette quantization. The overall error metric shows 7.48\% error across rendered images.}
  \label{fig:appendix_cmc_quantization}
\end{figure}

\paragraph{Why quantization helps.}
The 2D segmentation maps provide a discrete set of valid semantic colors. However,
after deblending, a Gaussian color $\hat{s}_i$ may lie between two or more valid
colors because of alpha blending. Palette quantization corrects this by snapping
the recovered color back to the closest valid segmentation color (Figure~\ref{fig:appendix_cmc_quantization}):
\begin{equation}
s_i^{q}
=
{\rm arg\,min}_{c\in\mathcal{P}}
d_{\mathrm{CMC}}
\left(
\mathrm{Lab}(\hat{s}_i),
\mathrm{Lab}(c)
\right).
\label{eq:appendix_overlap_quantization}
\end{equation}
This step does not remove the fact that overlap happened. Instead, it uses the
discrete structure of the 2D segmentation labels to make the final Gaussian map
semantically cleaner. The deblending step estimates where the Gaussian color lies
from all rendered observations, and the quantization step maps this estimate back
to a valid object-level label.

\section{Further Experiments}
\label{app:FEX}

In this appendix, we provide additional experiments and observations about the
sensitivity of our method to 2D segmentation quality. Since our pipeline assigns
semantics to 3D Gaussians from 2D segmentation evidence, the quality of the input
segmentations directly affects the quality of the final semantic Gaussian map.
These experiments are not meant to evaluate a specific segmentation model.
Instead, they show what types of segmentation errors are harmful, why they affect
the deblending and quantization stages, and what practical conditions make the
method work reliably.

\paragraph{Failure case from segmentation noise.}
\label{app:segmentation_noise_failure}

\begin{figure}[h]
  \centering
  \includegraphics[width=\linewidth]{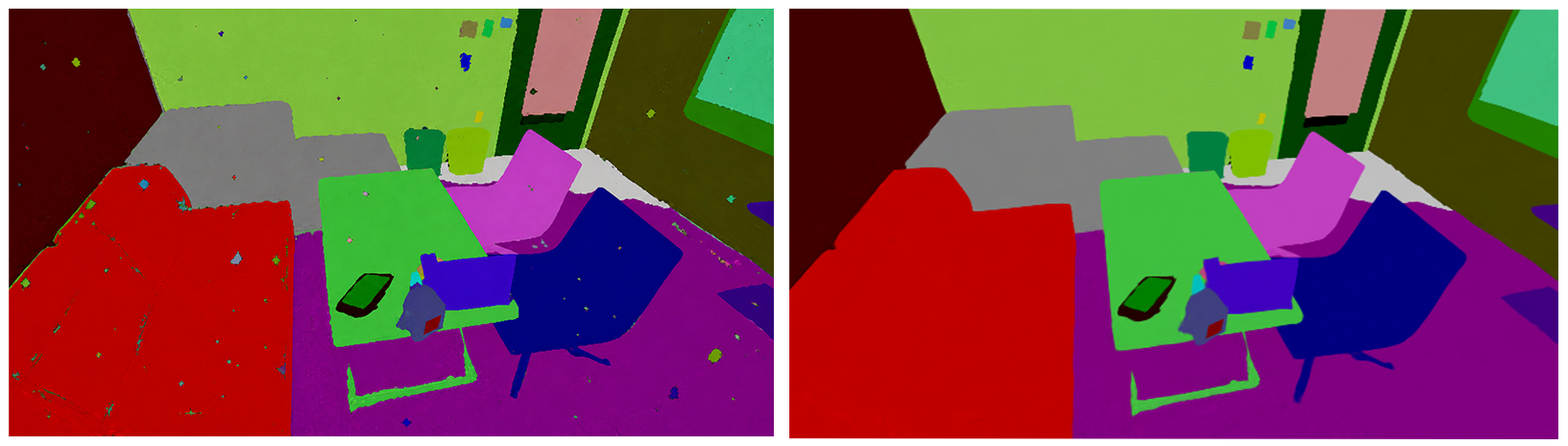}
  \caption{
  Failure case caused by noisy 2D segmentation inputs.
  Left: semantic rendering when the input segmentation contains artificial noise
  and segmentation artifacts. Right: semantic rendering obtained from clean
  segmentation inputs. Since our method uses 2D segmentation colors as semantic
  evidence for the Gaussians, noisy labels can be accumulated during deblending
  and can lead to incorrect semantic assignments in the final 3D map.
  }
  \label{fig:segmentation_noise_failure}
\end{figure}

Figure~\ref{fig:segmentation_noise_failure} shows a representative failure case
caused by noisy 2D segmentations. The right image shows the semantic rendering
obtained from clean segmentation inputs, while the left image shows the result
when the input segmentations are corrupted by artificial noise and segmentation
artifacts. The noisy input produces a less consistent semantic rendering from the
3D Gaussian map.

This happens because our method relies on the 2D segmentation maps as semantic
observations. During deblending, each Gaussian collects color evidence from all
pixels that it helps render. If the segmentation color at a pixel is wrong, then
that incorrect color is also accumulated as evidence for the Gaussians that
contributed to that pixel. When this happens repeatedly across frames, the
estimated Gaussian semantic color can be pulled away from the correct object
color. The problem becomes more visible near object boundaries, where a Gaussian
may already contribute to multiple neighboring regions due to alpha blending.

Therefore, noisy segmentation does not only create local 2D errors. It can also
propagate into the 3D map because the same Gaussian is observed in multiple
frames. If a wrong semantic label appears consistently, or if boundary noise is
present in many views, the final assigned color of a Gaussian may become
incorrect. This explains why segmentation quality is an important practical
requirement for our pipeline.

It is important to note that our method is not tied to a particular segmentation
model. The user can choose any segmentation method as long as the resulting
segmentation maps are reliable enough. In practice, better segmentation masks
lead to cleaner semantic Gaussian assignments, while noisy masks naturally lead
to noisier 3D semantics.

\paragraph{Color-confusable object experiment.}
\label{app:color_confusable_experiment}

We also evaluate a more challenging setting where different object instances are
assigned visually similar segmentation colors. This experiment is designed to
test the palette quantization stage. Since palette quantization maps each
deblended Gaussian color to the closest valid color in the segmentation palette,
it can become difficult when two valid colors are very close in the chosen color
space.

\begin{figure}[h]
  \centering
  \includegraphics[width=\linewidth]{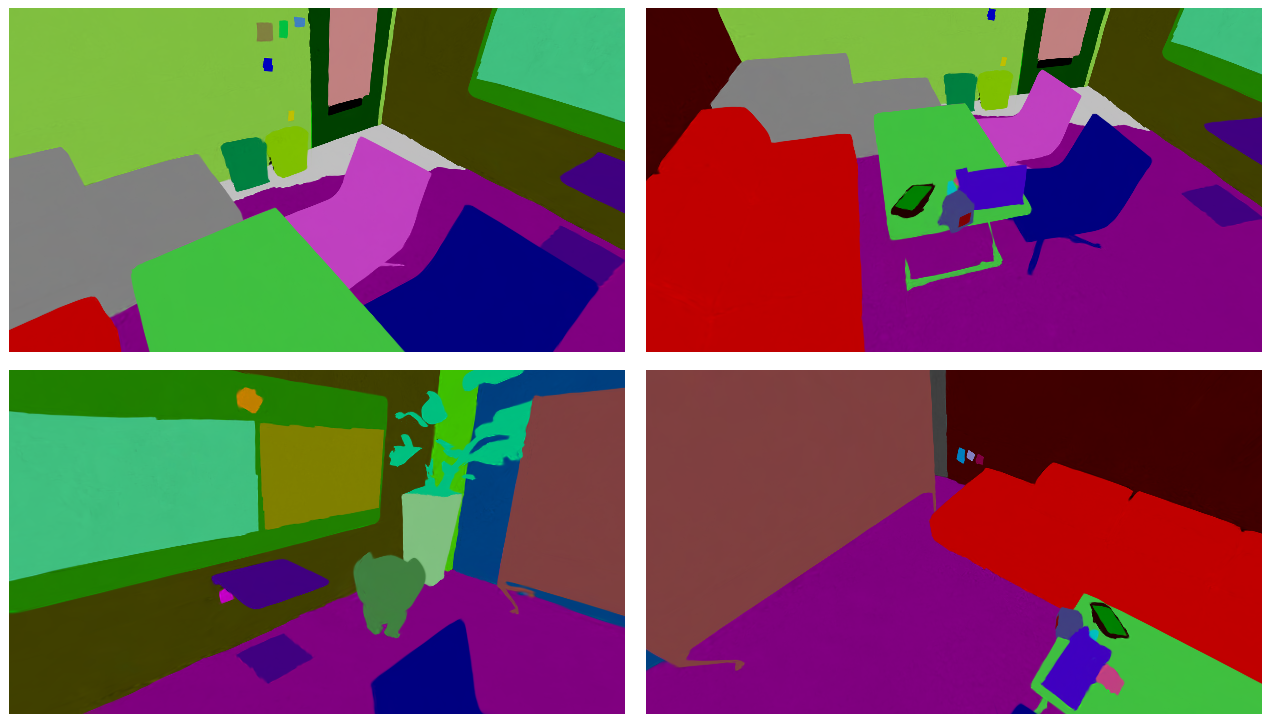}
  \caption{
  Example of color-confusable segmentation inputs. Different objects or object
  parts are assigned similar colors, making the palette quantization step more
  challenging. The experiment tests whether the recovered Gaussian colors can
  still be mapped back to the correct discrete semantic colors.
  }
  \label{fig:color_confusable_palette}
\end{figure}

\begin{figure}[h]
  \centering
  \includegraphics[]{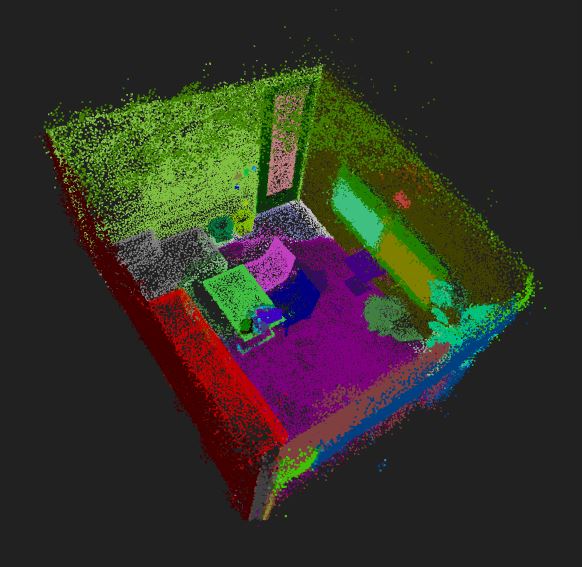}
  \caption{
  Semantic rendering result for the color-confusable experiment. The method
  remains reliable when similar colors are still sufficiently separated in the
  selected color space. When colors become too close, however, palette
  quantization can become ambiguous.
  }
  \label{fig:color_confusable_result}
\end{figure}

Figure~\ref{fig:color_confusable_palette} shows examples of similar colors used
for different parts of the scene, and Figure~\ref{fig:color_confusable_result}
shows the corresponding semantic Gaussian rendering. In our experiments, the
method remains stable even when the colors are close, as long as the color
distance is still large enough for the quantization metric to separate them. This
is because the deblending stage aggregates evidence over many pixels and frames,
and the quantization stage then snaps the estimated color to one of the valid
palette entries.

However, this experiment also highlights an important limitation. If two object
colors are almost identical in the quantization space, then palette quantization
can fail because the nearest-color decision becomes ambiguous. In such cases, a
Gaussian whose recovered color lies between two very similar palette entries may
be assigned to the wrong object. This is not a limitation of a specific
implementation, but a natural ambiguity caused by using colors as object
identifiers but still our method can handle them at an acceptable margin.

In our implementation, we use CMC distance in CIELAB space because it gives a
perceptually meaningful comparison between colors. Nevertheless, the choice of
quantization metric is modular. For special cases, the user may choose a
different color space or distance function if it better separates the specific
palette used by the segmentation method. In general, assigning more separated
colors to different objects makes the quantization stage more robust.

\paragraph{Ablation on segmentation quality.}
\label{app:segmentation_quality_ablation}

To better understand how segmentation quality affects the final semantic map, we
perform an ablation study by changing the quality of the 2D segmentation inputs.
The purpose of this ablation is to isolate the dependency of our method on the
semantic observations. The 3D Gaussian geometry, camera poses, rendering
procedure, and quantization method are kept fixed, while only the segmentation
inputs are changed.

We consider several segmentation quality settings. In the clean setting, the
original segmentation maps are used without additional corruption. In the
boundary-noise setting, errors are added around object boundaries, where Gaussian
overlap is already more likely. In the random-label-noise setting, a percentage
of pixels is randomly assigned to incorrect palette colors. In the mixed-boundary
setting, boundary pixels are replaced by intermediate RGB values between
neighboring object colors. Finally, in the region-artifact setting, small holes,
thin noisy structures, or isolated mislabeled regions are added to simulate
typical segmentation artifacts.

The results show that clean segmentations produce the most consistent semantic
Gaussian maps. Boundary noise and mixed boundary colors are especially harmful
because they occur in the same regions where Gaussian footprints naturally
overlap. Random label noise also degrades the result, but its effect depends on
how often the corrupted labels are observed and accumulated across frames.
Region artifacts can also create incorrect local assignments, especially when
small noisy structures are repeatedly visible from multiple viewpoints.

This ablation suggests three practical rules of thumb for preparing segmentation
inputs for our method. First, segmentation masks should avoid mixed RGB values at
object boundaries. Some segmentation pipelines produce smooth transitions between
two object colors near boundaries, but this is harmful for our method because
those intermediate colors do not correspond to valid object labels. Second,
object colors should be well separated in the color space used by the
quantization algorithm. Larger distances between palette colors reduce the chance
that a blended or noisy Gaussian color is snapped to the wrong object. Third,
segmentation masks should avoid sharp isolated artifacts, such as thin noisy
structures or small mislabeled regions, because these can introduce incorrect
semantic evidence during the deblending stage.

\paragraph{Artificial corruption of 2D segmentations.}
\label{app:artificial_segmentation_corruption}

Finally, we artificially corrupt the 2D segmentation maps and measure the drop in
semantic quality. Starting from the clean segmentation maps, we add increasing
levels of noise and then run the same semantic assignment pipeline. This allows
us to measure how robust the method is when the segmentation input becomes less
reliable.

Let $\mathrm{mIoU}_{\mathrm{clean}}$ be the result obtained using the original
segmentation maps, and let $\mathrm{mIoU}_{\mathrm{noise}}$ be the result after
corrupting the segmentations. We report the performance drop as
\begin{equation}
\Delta \mathrm{mIoU}
=
\mathrm{mIoU}_{\mathrm{clean}}
-
\mathrm{mIoU}_{\mathrm{noise}}.
\label{eq:miou_drop}
\end{equation}

We corrupt the segmentation maps using random pixel-level label noise. For a
noise level $\rho$, a fraction $\rho$ of pixels is randomly selected and replaced
with another color from the valid segmentation palette. This keeps the corrupted
image inside the same palette, but breaks the spatial consistency of the labels.
We also test boundary-focused corruption, where noise is applied mostly around
object boundaries. This setting is more challenging because boundary pixels are
the most likely to influence multiple Gaussians through alpha blending.

After corruption, the final semantic rendering quality decreases compared with
the clean segmentation setting. The drop in mIoU becomes larger as the corruption
level increases. Boundary-focused corruption causes a stronger drop than uniform
random noise, because boundary regions already contain larger alpha-compositing
ambiguity. In other words, errors near object edges are more likely to affect the
semantic color assigned to nearby Gaussians.

These results confirm that corrupted 2D segmentations lead to weaker final
semantic maps. This is expected because the 2D segmentation maps are the semantic
source used by the pipeline. When the input labels are corrupted, the deblending
stage receives incorrect observations, and the quantization stage may snap some
Gaussians to wrong palette colors.

Taken together, these experiments show where the method works well and where it
needs cleaner input. The pipeline can handle moderate noise and visually close
palette colors when the masks are consistent and the object colors remain
separable. However, errors that repeat across frames, especially near object
boundaries, can be accumulated by the Gaussians and appear again in the final 3D
semantic map. For this reason, the segmentation stage should be treated as an
important input preparation step rather than a separate detail of the pipeline.

\section{Broader Impact}
\label{app:broader_impact}

This work is primarily methodological and operates on monocular videos provided
by the user. Therefore, its societal impact depends strongly on the input data
and deployment context. Potential positive uses include lightweight 3D scene
understanding for robotics, augmented reality, inspection, and assistive systems.
However, if applied to sensitive indoor spaces without consent, the same
capability could raise privacy concerns or support unauthorized spatial mapping.
Practical deployments should therefore use consented data, avoid privacy-sensitive
scenes when possible, and follow local privacy and data-protection requirements.
We do not release any new scraped dataset or high-risk generative model; any
released code is intended to reproduce the proposed semantic mapping pipeline on
consented or public benchmark data.



\end{document}